\documentclass[11pt,a4paper]{article}

\usepackage[utf8]{inputenc}
\usepackage[T1]{fontenc}
\usepackage{lmodern}
\usepackage{microtype}

\usepackage[margin=1in]{geometry}

\usepackage{amsmath,amssymb}

\usepackage{graphicx}
\graphicspath{{figures/}}
\usepackage[font=small,labelfont=bf]{caption}
\usepackage{subcaption}

\usepackage{booktabs}
\usepackage{multirow}
\usepackage{array}
\newcolumntype{C}[1]{>{\centering\arraybackslash}p{#1}}

\usepackage[dvipsnames]{xcolor}
\usepackage[colorlinks=true,linkcolor=MidnightBlue,citecolor=OliveGreen,urlcolor=BrickRed]{hyperref}

\usepackage[numbers,sort&compress]{natbib}

\usepackage{enumitem}
\usepackage{url}
\usepackage{float}
\usepackage{pifont}

\title{\textbf{Criterion Validity of LLM-as-Judge\\
for Business Outcomes in Conversational Commerce}}

\author{
  Liang Chen, Qi Liu, Wenhuan Lin, Feng Liang
}
\date{March 2026}

\begin{document}
\maketitle

\begin{abstract}
Multi-dimensional rubric-based dialogue evaluation is widely used to assess conversational AI, yet its \emph{criterion validity}---whether quality scores predict the downstream outcomes they are meant to serve---remains largely untested. We address this gap through a two-phase study on a major Chinese matchmaking platform, testing the criterion validity of a 7-dimension evaluation rubric (implemented via LLM-as-Judge) against verified business conversion. Our findings concern the rubric's dimension design and weighting, not the LLM judge's scoring accuracy per se: any judge---human or LLM---using the same rubric would face the same structural issue. Our core finding is \textbf{dimension-level heterogeneity}: quality dimensions differ dramatically in their association with conversion. In the expanded Phase~2 ($n = 60$ human conversations, stratified random sample, verified conversion labels), Need Elicitation (D1: $\rho = 0.368$, $p = 0.004$, Cohen's $d = 0.74$) and Pacing Strategy (D3: $\rho = 0.354$, $p = 0.006$, $d = 0.77$) are significantly associated with conversion after Bonferroni correction, while Contextual Memory (D5: $\rho = 0.018$, n.s.) shows no detectable association. This heterogeneity causes the equal-weighted composite ($\rho = 0.272$) to underperform its best individual dimensions---a structural \emph{composite dilution} effect that conversion-informed reweighting (D3 = 40\%, D5 = 0\%) partially corrects ($\rho = 0.351$, $p = 0.006$). Logistic regression controlling for conversation length confirms that D3's association with conversion strengthens (OR $= 3.18$, $p = 0.006$) rather than attenuates, ruling out a conversation-length confound. An initial pilot ($n = 14$) that mixed human and AI conversations had produced a misleading ``evaluation-outcome paradox'' (higher scores associated with worse outcomes), which Phase~2 revealed to be largely an artifact of the agent-type confound rather than a fundamental failure of quality scoring. Complementary behavioral analysis of 130 conversations through a Trust-Funnel framework identifies a candidate mechanism: the AI executes sales behaviors without building user trust (72\% reach closing yet 0\% reach trust threshold). We operationalize these findings in a three-layer evaluation architecture (L3~Safety $\to$ L2~Quality $\to$ L1~Business) deployed through two iterative evaluation cycles, and advocate criterion validity testing as standard practice in applied dialogue evaluation.
\end{abstract}


\section{Introduction}
\label{sec:intro}

\subsection{The Criterion Validity Question}
\label{sec:criterion}

The deployment of large language models (LLMs) as conversational agents in commercial settings has created an urgent need for evaluation frameworks that reliably assess agent performance. Multi-dimensional quality scoring---rating dialogues across dimensions such as coherence, empathy, relevance, and informativeness---has become the dominant paradigm, reinforced by the rise of LLM-as-Judge approaches~\citep{liu2023geval,zheng2023mtbench,li2024survey}. These frameworks assume, implicitly or explicitly, that higher quality scores indicate better system performance.

But do they? In psychometric terms, this is a question of \emph{criterion validity}---the degree to which a measurement instrument predicts an external criterion of interest~\citep{messick1995}. While the dialogue evaluation community has invested heavily in establishing \emph{construct validity} (does the metric measure what it claims to measure?) and \emph{inter-rater reliability} (do judges agree?), the question of whether quality scores predict the outcomes they are meant to serve remains largely unexamined.

This omission matters because the gap between measurement and outcome is well-theorized across multiple disciplines. Goodhart's Law~\citep{goodhart1975} warns that when a measure becomes a target, it ceases to be a good measure. The satisfaction-loyalty gap in marketing research demonstrates that customer satisfaction scores do not reliably predict purchase behavior~\citep{oliver1999}. The surrogate endpoint problem in clinical trials shows that biomarkers can improve while patient outcomes worsen~\citep{fleming1996}. In machine learning, the proxy alignment problem---where optimizing a proxy objective diverges from the true objective---has become a central concern in AI safety research~\citep{amodei2016}.

Given these well-established precedents, the absence of criterion validity testing in dialogue evaluation is not merely a gap---it is a risk. If quality scores are not associated with outcomes, systems optimized on these scores may improve on metrics while degrading on the objectives that matter.

\paragraph{Scope clarification.} We test the criterion validity of a multi-dimensional evaluation \emph{rubric}---specifically, whether its dimensions and their weighting are associated with business conversion. The rubric is implemented via an LLM judge (Claude Opus 4.6), but our findings are about rubric design, not judge accuracy: D5 (Contextual Memory) shows no association with conversion because the dimension itself is irrelevant to the outcome, not because the LLM mismeasures it. This distinction is important because the intervention implied by our findings is \emph{rubric redesign} (reweighting dimensions based on outcome associations), not \emph{judge replacement}. We use ``LLM-as-Judge'' in the title because rubric-based multi-dimensional scoring is the dominant paradigm within that ecosystem, and our methodology is proposed specifically for practitioners in that context.

\subsection{Research Context}
\label{sec:context}

We investigate this question through a case study from a major Chinese matchmaking platform, where an LLM-based AI agent was deployed to conduct sales conversations with parents seeking matchmaking services for their adult children. This context represents what we term \emph{high-emotion B2C sales}: transactions involving deeply personal life decisions, significant financial commitment (\textyen469--\textyen1,688), and trust relationships that must be built through intimate personal disclosure before any commercial discussion can succeed.

The research proceeded in two phases. A pilot study ($n = 14$, purposive sampling of mixed human and AI conversations) suggested that quality scores and conversion outcomes might be misaligned, motivating a systematic investigation. An expanded study ($n = 60$, stratified random sample of human conversations with verified conversion labels) then quantified the core phenomenon: individual quality dimensions exhibit dramatically different associations with business conversion, and equal-weighted composites are suboptimal because they dilute conversion-predictive dimensions with non-predictive ones. We further operationalized these findings in a three-layer evaluation system deployed through two iterative cycles, demonstrating the full path from criterion validity diagnosis to evaluation redesign.

\subsection{Theoretical Framework}
\label{sec:theory}

Our investigation is grounded in three theoretical streams:

\paragraph{Criterion validity in evaluation design.} Walker et al.'s PARADISE framework~\citep{walker1997} established that dialogue evaluation should include task success as a component, and demonstrated that different dialogue features have different predictive power for user satisfaction. Liu et al.~\citep{liu2016} subsequently showed that standard automatic metrics (BLEU, METEOR) have near-zero correlation with human judgments. We extend this line of inquiry from \emph{user satisfaction} to \emph{business outcomes}, asking whether human-designed multi-dimensional quality scores fare better than automatic metrics in predicting downstream success.

\paragraph{Proxy metric failure.} Goodhart's Law and its formalization in AI alignment research~\citep{amodei2016} predict that proxy metrics can systematically diverge from true objectives, particularly when (a)~the proxy captures only a subset of the factors driving the outcome, and (b)~the system being evaluated excels on the non-predictive subset. Our multi-dimensional evaluation creates exactly this scenario: if some dimensions predict conversion while others do not (or predict negatively), a composite score can be misleading.

\paragraph{Trust-in-automation theory.} Lee \& See~\citep{lee2004} established that trust calibration---alignment between a user's trust and the system's trustworthiness---determines appropriate reliance. Hoff \& Bashir~\citep{hoff2015} extended this with dispositional, situational, and learned trust layers. In our context, trust calibration provides the theoretical mechanism linking dialogue quality to conversion: quality dimensions that build calibrated trust should predict conversion, while dimensions that inflate surface-level performance without building trust should not.

\subsection{Research Questions and Hypotheses}
\label{sec:rqs}

We investigate three research questions, each with a directional expectation framed as a hypothesis. These hypotheses serve to structure analysis; our two-phase design (pilot $n = 14$, expanded $n = 60$) provides increasing evidential strength but remains exploratory:

\smallskip\noindent\textbf{RQ1:} Are standard multi-dimensional dialogue quality scores associated with business conversion?\\
\emph{H1}: The composite quality score has suboptimal criterion validity---its association with conversion is weaker than that of its best individual dimensions---because individual dimensions have heterogeneous relationships with the outcome.

\smallskip\noindent\textbf{RQ2:} Which specific quality dimensions are most strongly associated with conversion, and what is the candidate mechanism?\\
\emph{H2}: Dimensions capturing strategic judgment (e.g., pacing, timing) are more strongly associated with conversion than dimensions capturing execution quality (e.g., memory, fluency), because conversion in high-emotion sales depends on trust-building strategy rather than information processing capability.

\smallskip\noindent\textbf{RQ3:} Can the association between quality scores and conversion be improved through empirically informed dimension reweighting?\\
\emph{H3}: A weight scheme that upweights conversion-associated dimensions and downweights non-associated ones will produce a composite score better aligned with business outcomes in-sample. (Note: out-of-sample validation is required to establish generalizability; see Section~\ref{sec:circular}.)

\subsection{Contributions}
\label{sec:contributions}

This paper makes four contributions:

\begin{enumerate}[leftmargin=*,itemsep=2pt]
  \item \textbf{Empirical evidence for dimension-level heterogeneity in criterion validity}---to our knowledge, the first study to systematically test whether multi-dimensional dialogue quality scores predict business conversion in a commercial deployment. Through a two-phase design (pilot $n = 14$, expanded $n = 60$ with verified conversion labels), we quantify the \emph{composite dilution effect}: equal-weighted composites ($\rho = 0.272$) underperform their best individual dimensions ($\rho = 0.368$) because non-predictive dimensions attenuate the signal from predictive ones.
  \item \textbf{A criterion validity testing methodology} applicable to any multi-dimensional evaluation: dimension-outcome association analysis $\to$ conversion-informed reweighting $\to$ independent validation. We demonstrate that this simple procedure improves criterion validity from $\rho = 0.272$ to $\rho = 0.351$ ($p = 0.006$).
  \item \textbf{A Trust-Funnel dual-track framework} distinguishing sales \emph{behavior} (Funnel stages F1--F6) from user \emph{trust states} (Trust Ladder T0--T5), providing the behavioral mechanism explaining why the AI executes sales correctly yet fails on outcomes.
  \item \textbf{A three-layer evaluation architecture} (L3~Safety $\to$ L2~Quality $\to$ L1~Business) operationalized in a deployed system with two iterative evaluation cycles, demonstrating the full path from criterion validity diagnosis to evaluation redesign.
\end{enumerate}

\subsection{Paper Organization}
\label{sec:org}

Section~\ref{sec:related} reviews related work. Section~\ref{sec:method} describes the research context, data, and methodology. Section~\ref{sec:validity} presents the criterion validity analysis---the core empirical contribution. Section~\ref{sec:architecture} introduces the three-layer architecture and conversion-informed weighting. Section~\ref{sec:trustfunnel} details the Trust-Funnel framework. Section~\ref{sec:flywheel} presents the Failure-Driven Optimization Flywheel. Section~\ref{sec:discussion} discusses implications. Section~\ref{sec:limitations} addresses limitations and future work. Section~\ref{sec:conclusion} concludes.

\section{Related Work}
\label{sec:related}

\subsection{Dialogue Evaluation: Quality Metrics and Their Validity}

The evaluation of dialogue systems has evolved through four paradigmatic shifts. \textbf{Reference-based metrics} (BLEU, METEOR) dominated early work but were debunked by Liu et al.~\citep{liu2016}, who showed near-zero correlation (Spearman $\rho < 0.15$) with human judgments---an early signal that evaluation metrics may lack criterion validity. \textbf{Learned evaluation models} emerged as alternatives: USR~\citep{mehri2020usr} proposed multi-dimensional reference-free evaluation; FED~\citep{mehri2020fed} leveraged language model probabilities as quality signals.

The third paradigm---\textbf{LLM-as-Judge}---was catalyzed by G-Eval~\citep{liu2023geval}, demonstrating GPT-4's evaluation capability with chain-of-thought guidance (Spearman $\rho = 0.514$ with humans on summarization). MT-Bench~\citep{zheng2023mtbench} established the strong-model-evaluates-weak-model standard. Comprehensive surveys~\citep{li2024survey,li2024generation} cataloged systematic biases: verbosity preference, self-enhancement, and position bias. Multi-agent judge systems~\citep{sedoc2024} address single-evaluator bias through deliberation. EvalLM~\citep{kim2024evallm} introduces interactive criteria co-creation. Guan et al.~\citep{guan2025} synthesize nearly 250 works on multi-turn evaluation. Most recently, the \textbf{crowd-preference} paradigm---exemplified by Chatbot Arena's ELO-based ranking from large-scale human pairwise votes~\citep{zheng2023mtbench}---has emerged as a de facto standard for model comparison, and underpins the preference data used in RLHF and DPO training~\citep{rafailov2023}.

\paragraph{A shared validation anchor.} Critically, all four paradigms---reference-based, learned, LLM-as-Judge, and crowd-preference---share a common validation anchor: \emph{human judgment}. Reference-based metrics are validated by correlation with human ratings; learned models are trained or evaluated against human annotations; LLM-as-Judge systems are benchmarked by agreement with human scores; and crowd-preference platforms (e.g., Chatbot Arena) use human votes directly. ``Quality'' in dialogue evaluation has thus been operationally synonymous with ``agreement with human preferences.'' The question of whether human preferences themselves predict downstream behavioral outcomes---whether what humans \emph{judge} as good dialogue actually \emph{produces} good outcomes---has remained outside the evaluation paradigm entirely. This is not an accidental omission but a structural feature of how the field defines evaluation success.

\paragraph{The criterion validity gap.} Across this extensive literature, evaluation quality is assessed through inter-rater agreement, construct coverage, and correlation with human preferences. What is largely absent is validation against \emph{external outcomes}. Walker et al.~\citep{walker1997} tested dialogue features against user satisfaction---the closest precedent---but user satisfaction is itself a subjective judgment, not a behavioral outcome. To our knowledge, no existing work has tested whether multi-dimensional dialogue quality scores are associated with business conversion or other objectively verifiable downstream outcomes. Our work provides an initial, exploratory examination of this gap.

\subsection{Proxy Metrics and the Evaluation-Outcome Gap}

Walker et al.'s \textbf{PARADISE framework}~\citep{walker1997} is the closest precedent in dialogue evaluation. PARADISE used linear regression to identify which dialogue features predict user satisfaction, establishing the principle that not all measurable features are equally predictive. Our work extends PARADISE's approach from user satisfaction to business conversion, and from feature-level to dimension-level analysis within a multi-dimensional scoring rubric.

\textbf{Goodhart's Law}~\citep{goodhart1975} and Campbell's Law~\citep{campbell1979} formalize the intuition that optimizing a proxy metric can corrupt the metric's predictive value. In AI alignment, this has been operationalized as the \emph{reward hacking} problem~\citep{amodei2016}, where agents learn to maximize a reward signal without achieving the intended objective.

The \textbf{satisfaction-loyalty gap} in marketing research~\citep{oliver1999,reichheld2003} demonstrates that satisfied customers do not necessarily become loyal or repeat purchasers---a direct analogy to our finding that high-quality dialogues do not necessarily produce conversions.

In clinical trials, the \textbf{surrogate endpoint problem}~\citep{fleming1996} has led to cases where treatments improved biomarkers while worsening patient outcomes. The structural parallel to dialogue evaluation is precise: individual quality dimensions are surrogate endpoints, and business conversion is the clinical outcome.

\subsection{AI Sales, Persuasion, and Trust}

Luo et al.~\citep{luo2019} conducted the first large-scale experiment on AI sales, finding that AI identity disclosure reduced purchase rates by over 79\%. Salvi et al.~\citep{salvi2024} demonstrated that GPT-4 matches human persuasive ability with strategic personalization. On the systems side, AI-Salesman~\citep{zhang2025salesman} introduces reinforcement learning for telemarketing; CSales~\citep{kim2025csales} unifies preference elicitation and persuasion; SalesRLAgent~\citep{nandakishor2025} models sales as sequential decision problems.

\textbf{Trust-in-automation theory} is central to our analysis. Lee \& See~\citep{lee2004} established the three-basis trust model: performance, process, and purpose. Hancock et al.~\citep{hancock2011} meta-analyzed human-robot trust and found that robot performance characteristics explain more variance than human or environmental factors. In our context, this predicts that the AI's \emph{strategic performance} (pacing, timing) should matter more for trust-building than its \emph{technical capabilities} (memory, fluency)---precisely what our data shows.

\subsection{Positioning This Work}

\begin{figure}[t]
  \centering
  \includegraphics[width=0.65\columnwidth]{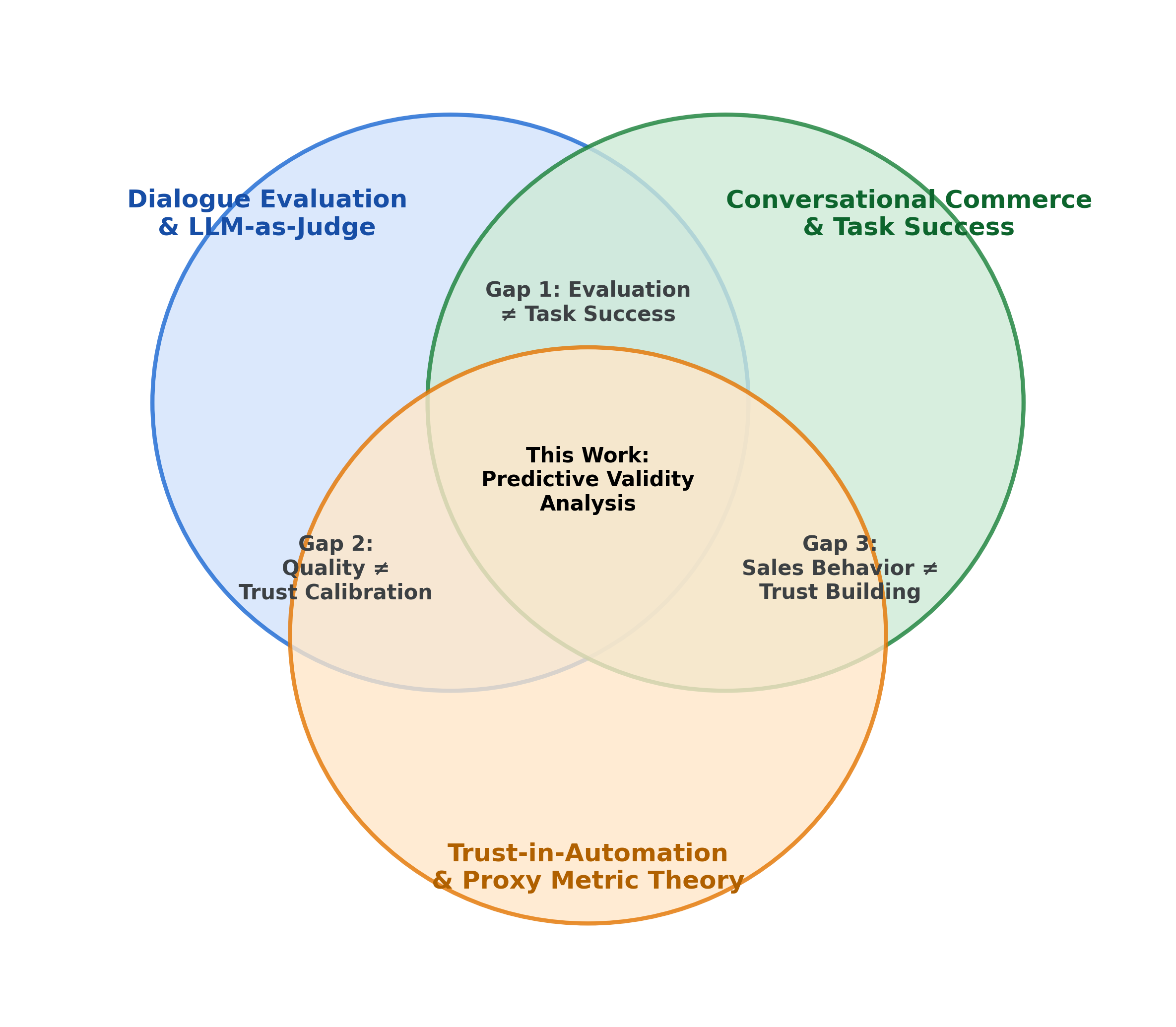}
  \caption{Theoretical positioning. This work sits at the intersection of dialogue evaluation (LLM-as-Judge), conversational commerce (task success), and trust-in-automation \& proxy metric theory, addressing three research gaps.}
  \label{fig:venn}
\end{figure}

Our research sits at the intersection of dialogue evaluation methodology, proxy metric theory, and AI-mediated commerce (Figure~\ref{fig:venn}). The primary contribution is methodological: we challenge the implicit assumption that human-preference-validated quality scores are sufficient for applied deployment, and bring criterion validity analysis---standard practice in psychometrics and clinical research---to dialogue evaluation. Where the existing paradigm asks ``does the metric agree with human judgment?''\ (construct validity), we ask ``does the metric predict the outcome it is meant to serve?''\ (criterion validity). We provide initial exploratory evidence for a failure mode that existing theory predicts but that, to our knowledge, no prior work has empirically examined in this domain.

\section{Research Context and Methodology}
\label{sec:method}

\subsection{Platform and Deployment Context}
\label{sec:platform}

Our study is situated on a major Chinese matchmaking platform (anonymized) with tens of millions of registered users. The platform operates a ``parent matchmaking'' service where parents register on behalf of their adult children and interact with consultants to find potential partners. In 2025, the platform deployed an LLM-based AI matchmaking consultant to augment human agents, operating through WeChat Work.

\paragraph{Service pricing.} \textyen469/month, \textyen899/3 months, or \textyen1,688/year for premium matchmaking services.

\paragraph{Conversion operationalization.} The two study phases use different conversion criteria, reflecting data access improvements. \emph{Phase~1} uses Trust Ladder stage T5 (``Price Reasonable''---user actively engages in pricing discussion with purchase intent) as a proxy for conversion, because payment system access was unavailable during the pilot. In the annotated data, all conversations reaching T5 resulted in completed payment (3/3 cases), though this perfect alignment may not hold in larger samples. \emph{Phase~2} uses verified binary conversion labels (\emph{is\_converted}) from operational records, directly reflecting completed payment transactions. This hard outcome measure eliminates the proxy inference chain. All core criterion validity findings (Section~\ref{sec:h2} onward) are based on Phase~2's verified labels. We discuss remaining limitations of both criteria in Section~\ref{sec:proxy}.

\subsection{Data Collection and Sampling}
\label{sec:sampling}

Data collection proceeded in two phases (Table~\ref{tab:datasets}).

\begin{table}[t]
  \centering
  \small
  \caption{Dataset summary across two study phases.}
  \label{tab:datasets}
  \begin{tabular}{llcl}
    \toprule
    \textbf{Dataset} & \textbf{Source} & \textbf{Size} & \textbf{Purpose} \\
    \midrule
    \multicolumn{4}{l}{\emph{Phase 1 (Pilot)}} \\
    Golden Conv. & Human agent export & 170 (30 annotated) & Trust Ladder annotation \\
    AI Conv. & PolarDB query & 202 (100+102) & Funnel analysis \\
    Judge-Scored & Purposive sample & 15 (5H + 10AI) & Quality evaluation \\
    Criterion Validity & Scored $-$ T6 case & 14 & Pilot correlation \\
    \midrule
    \multicolumn{4}{l}{\emph{Phase 2 (Expanded)}} \\
    Human Conv. Pool & WeChat Work logs & 59,316 (319 conv.) & Sampling frame \\
    Expanded Sample & Stratified random & 60 (25C + 35NC) & Criterion validity \\
    \bottomrule
  \end{tabular}
\end{table}

\paragraph{Phase 1: Purposive pilot ($n = 14$).} The Phase~1 Judge-Scored Set was not randomly sampled. The 5 human conversations were drawn from completed transactions, biasing the human sample toward successful outcomes. The 10 AI conversations were selected for high engagement, not for conversion outcomes. This means: (a)~the human-vs-AI comparison (60\% vs.\ 0\% conversion) reflects sampling bias as much as performance differences; (b)~the Phase~1 criterion validity analysis focuses on the \emph{within-sample} question: regardless of agent type, do quality dimension scores predict which conversations reach conversion?

\paragraph{Phase 2: Stratified random sample ($n = 60$).} To address Phase~1's limitations---small sample size, purposive sampling, and human-AI confounding---we expanded the criterion validity analysis using a stratified random sample drawn exclusively from human agent conversations with verified conversion labels. The sampling frame comprised 59,316 conversations from WeChat Work operational logs (March--July 2025), of which 319 had verified conversion (\emph{is\_converted} = 1). We sampled 25 converted conversations ($\geq 5$ user messages, randomly drawn from 170 golden conversations, seed = 42) and 35 unconverted conversations ($\geq 8$ user messages to ensure substantive interaction, stratified across agents). All 60 conversations were scored using the identical v2.0 LLM-as-Judge rubric (Section~\ref{sec:judge}). The population base rate is very low (${\sim}0.5\%$ conversion, 319/59{,}316); the 25:35 ratio substantially oversamples converted cases to ensure adequate statistical power for between-group comparisons. This stratified oversampling is standard practice for rare-event studies but means that sample proportions do not reflect population prevalence. By restricting to human conversations, Phase~2 eliminates the agent-type confound that pervaded Phase~1.

\subsection{LLM-as-Judge Design}
\label{sec:judge}

The evaluation rubric comprises 7 dimensions on a 1--5 scale (Table~\ref{tab:rubric}).

\begin{table}[t]
  \centering
  \small
  \caption{Evaluation rubric: 7 quality dimensions.}
  \label{tab:rubric}
  \begin{tabular}{llp{5.5cm}c}
    \toprule
    \textbf{Dim.} & \textbf{Code} & \textbf{Description} & \textbf{v2.0 Wt.} \\
    \midrule
    Need Elicitation & D1 & Completeness of user need discovery & 20\% \\
    Emotional Empathy & D2 & Quality of emotional recognition \& response & 20\% \\
    Pacing Strategy & D3 & Strategic judgment of when to advance/pause & 20\% \\
    Objection Handling & D4 & Effectiveness at addressing user concerns & 15\% \\
    Contextual Memory & D5 & Retention and use of conversation history & 10\% \\
    Product Accuracy & D6 & Correctness of product/pricing information & 10\% \\
    Brand Consistency & D7 & Alignment with platform standards & 5\% \\
    \bottomrule
  \end{tabular}
\end{table}

The v2.0 rubric incorporated four bias controls targeting known LLM-as-Judge biases~\citep{li2024survey}: (1)~verbosity bias, (2)~self-enhancement bias, (3)~surface fluency bias, and (4)~emotional over-weighting. The judge was implemented using Claude Opus 4.6 (Anthropic, 2025), temperature $= 0$, with chain-of-thought reasoning required before scoring each dimension. We acknowledge the self-evaluation risk and discuss mitigation in Section~\ref{sec:selfeval}.

\subsection{Trust Ladder Annotation}
\label{sec:trustladder}

Thirty golden conversations were stratified by duration and annotated with a six-stage Trust Ladder (Table~\ref{tab:trustladder}).

\begin{table}[t]
  \centering
  \small
  \caption{Trust Ladder stages with behavioral indicators.}
  \label{tab:trustladder}
  \begin{tabular}{clp{5cm}c}
    \toprule
    \textbf{Stage} & \textbf{Name} & \textbf{Behavioral Indicator} & \textbf{Avg.\ Msgs} \\
    \midrule
    T0 & No Trust & User ignores or rejects all contact & --- \\
    T1 & Platform Credible & User responds, begins basic interaction & 4.9 \\
    T2 & Agent Helpful & User proactively shares personal info & 12.6 \\
    T3 & Success Evidence & User believes platform can deliver & 10.8 \\
    T4 & My Child Can Match & User discusses specific requirements & 23.3 \\
    T5 & Price Reasonable & User enters purchase discussion & 36.4 \\
    T6 & Trust Collapse & Post-purchase trust breakdown & --- \\
    \bottomrule
  \end{tabular}
\end{table}

Annotation was performed by LLM (Claude) with 462 key turns identified across 30 conversations. We acknowledge the absence of human annotation validation as a limitation (Section~\ref{sec:trustval}).

\subsection{Behavioral Funnel Detection}

A rule-based detector classified each message into one of six sales behavior stages (F1 Rapport $\to$ F2 Need Elicitation $\to$ F3 Pain Point Activation $\to$ F4 Product Introduction $\to$ F5 Objection Handling $\to$ F6 Closing). Validated on 5 manually checked conversations with ${\sim}80\%$ accuracy for maximum stage identification.

\subsection{Statistical Methods}

Primary analyses use non-parametric methods: Spearman rank correlation ($\rho$) for dimension-conversion association, Cohen's $d$ for effect size estimation, and Bonferroni correction for multiple comparisons across 7 dimensions. We report Bonferroni-adjusted $p$-values (uncorrected $p$ multiplied by 7), which are compared against the standard $\alpha = 0.05$ threshold. For robustness, we supplement with logistic regression to test dimension--conversion associations while controlling for conversation length (message count), partial Spearman correlations, and 4-fold temporal cross-validation of the weight scheme (Section~\ref{sec:robustness} and~\ref{sec:circular}).

\paragraph{Power considerations.} Phase~1 ($n = 14$, deal group $n = 3$) has approximately 80\% power to detect only very large effects ($\rho \geq 0.65$). Phase~2 ($n = 60$, converted group $n = 25$) has approximately 80\% power to detect moderate effects ($\rho \geq 0.35$), providing substantially improved sensitivity while remaining underpowered for small effects.

\paragraph{Two-phase analysis strategy.} Phase~1 serves as hypothesis generation (which dimensions show potential association?). Phase~2 provides a larger, independently sampled test of these associations on human-only data with hard conversion labels. Phase~2 uses the same rubric but different conversations and a different sampling strategy, mitigating (though not eliminating) concerns about circular analysis.

\paragraph{Circular analysis caveat.} Within Phase~1, the same $n = 14$ dataset was used for discovering dimension-conversion correlations and optimizing the weight scheme---a form of double-dipping~\citep{kriegeskorte2009}. Phase~2 partially addresses this by testing the same dimensional hypotheses on independent data, but the weight scheme comparison in Phase~2 still uses the full dataset for both discovery and evaluation. We discuss remaining mitigation strategies in Section~\ref{sec:circular}.

\paragraph{Terminology note.} We use ``criterion validity'' throughout to refer to the degree to which evaluation scores are associated with an external criterion (business conversion). We avoid ``predictive validity'' (a subtype implying temporal precedence) because our analysis is concurrent rather than prospective: quality scores and conversion status were assessed on the same conversations. Future longitudinal studies would enable predictive validity claims.

\subsection{Ethical Considerations}

This study was conducted as an internal evaluation project using operational data under the platform's existing user agreement. All transcripts were de-identified: user names replaced with pseudonyms (H1--H5, A1--A10), phone numbers and location details redacted.

\paragraph{IRB status.} No formal Institutional Review Board (IRB) review was conducted for this study. We acknowledge this as a limitation. The study uses retrospective analysis of de-identified operational data with no user intervention, which may qualify for IRB exemption under many institutional policies (e.g., U.S.\ 45 CFR 46.104(d)(4) for secondary research on identifiable private information with adequate de-identification). However, we recognize that blanket ToS consent does not constitute informed consent for research publication, and that the involvement of a vulnerable population (elderly parents making financial decisions under emotional pressure) warrants heightened ethical scrutiny. We recommend formal ethics review for any future extensions involving prospective data collection, intervention studies (e.g., Trust Gate A/B testing), or cross-platform replication. We discuss broader ethical implications in Section~\ref{sec:ethics}.

\section{Criterion Validity Analysis}
\label{sec:validity}

This section presents the core empirical contribution: testing whether multi-dimensional quality scores are associated with conversion outcomes. We report results from both study phases: the pilot ($n = 14$, mixed human + AI, purposive) and the expanded analysis ($n = 60$, human-only, stratified random).

\subsection{Phase 1 Pilot: An Apparent Evaluation-Outcome Paradox ($n = 14$)}
\label{sec:h1}

Table~\ref{tab:scores} presents the v2.0 Judge scores for the 15 scored conversations in Phase~1.

\begin{table}[t]
  \centering
  \small
  \caption{Phase~1: v2.0 Judge scores---Human vs.\ AI agents.}
  \label{tab:scores}
  \begin{tabular}{lcccc}
    \toprule
    \textbf{Dimension} & \textbf{Wt.} & \textbf{Human $\mu$ (SD)} & \textbf{AI $\mu$ (SD)} & \textbf{Gap} \\
    \midrule
    D1 Need Elicitation & 20\% & 2.40 (0.89) & 3.00 (0.82) & +0.60 \\
    D2 Emotional Empathy & 20\% & 2.20 (0.84) & 2.60 (0.84) & +0.40 \\
    D3 Pacing Strategy & 20\% & 2.40 (0.55) & 1.60 (0.70) & $\mathbf{-0.80}$ \\
    D4 Objection Handling & 15\% & 2.40 (1.14) & 2.40 (0.97) & 0.00 \\
    D5 Contextual Memory & 10\% & 2.60 (0.89) & 3.70 (1.16) & $\mathbf{+1.10}$ \\
    D6 Product Accuracy & 10\% & 3.20 (1.10) & 3.10 (0.74) & $-0.10$ \\
    D7 Brand Consistency & 5\% & 2.60 (0.55) & 2.80 (0.79) & +0.20 \\
    \midrule
    \textbf{Weighted Total} & & \textbf{2.47} (0.76) & \textbf{2.62} (0.67) & $\mathbf{+0.15}$ \\
    \bottomrule
  \end{tabular}
\end{table}

All 15 conversations were scored (Table~\ref{tab:scores}); the criterion validity analysis uses $n = 14$ after excluding one trust collapse case (H2, coded T6) whose outcome is ambiguous. Among these 14, the no-deal group ($n = 11$) achieved a higher mean quality score than the deal group ($n = 3$): 2.70 vs.\ 2.48. The composite's Spearman correlation with Trust Ladder stage was weak and non-significant ($\rho = 0.355$, $p = 0.213$), indicating the composite score did not meaningfully distinguish conversion outcomes. Only D3 (Pacing Strategy) reached significance before correction ($\rho = 0.679$, $p = 0.008$), while D5 (Contextual Memory) showed a negative but non-significant association ($\rho = -0.284$, $d = -1.35$).

This pattern---AI scoring higher on quality yet achieving zero conversions---suggested an evaluation-outcome paradox. However, the pilot's design confounded agent type with conversion: all 3 deal cases were human, all 10 AI cases were no-deal. The apparent ``paradox'' could reflect genuine criterion validity failure, or it could be an artifact of comparing fundamentally different agent types with different sampling biases. Phase~2 was designed to disentangle these explanations.

\subsection{Phase 2: Dimension-Level Heterogeneity Confirmed ($n = 60$)}
\label{sec:h2}

Phase~2 scored 60 human-agent conversations (25 converted, 35 unconverted) with verified conversion labels, eliminating the agent-type confound. Table~\ref{tab:correlations} presents the full correlation analysis.

\begin{table}[t]
  \centering
  \small
  \caption{Phase~2: Dimension--conversion correlation analysis ($n = 60$, human conversations only). D4 and D5 have reduced $n$ due to inapplicable scoring (see text).}
  \label{tab:correlations}
  \begin{tabular}{lcccccc}
    \toprule
    \textbf{Dimension} & \textbf{$n$} & \textbf{Spearman $\rho$} & \textbf{$p$} & \textbf{Bonf.\ $p$} & \textbf{Cohen's $d$} & \textbf{Direction} \\
    \midrule
    D1 Need Elicit. & 60 & $\mathbf{+0.368}$ & $\mathbf{0.004}$ & $\mathbf{0.027^{*}}$ & $\mathbf{+0.74}$ & Conv $>$ Non-Conv \\
    D3 Pacing & 60 & $\mathbf{+0.354}$ & $\mathbf{0.006}$ & $\mathbf{0.039^{*}}$ & $\mathbf{+0.77}$ & Conv $>$ Non-Conv \\
    D2 Empathy & 60 & $+0.178$ & $0.174$ & $1.000$ & $+0.36$ & Conv $>$ Non-Conv \\
    D6 Product & 60 & $+0.156$ & $0.235$ & $1.000$ & $+0.35$ & Conv $>$ Non-Conv \\
    D4 Objection & 57 & $+0.143$ & $0.288$ & $1.000$ & $+0.32$ & Conv $>$ Non-Conv \\
    D7 Brand & 60 & $+0.138$ & $0.293$ & $1.000$ & $+0.27$ & Conv $>$ Non-Conv \\
    D5 Context.\ Memory & 58 & $+0.018$ & $0.895$ & $1.000$ & $-0.02$ & $\approx$ null \\
    \midrule
    \textbf{Composite (v2.0)}\footnotemark & $+0.272$ & $0.036$ & --- & $+0.56$ & \textbf{Conv $>$ Non-Conv} \\
    \bottomrule
  \end{tabular}
\end{table}
\footnotetext{The composite $\rho = 0.272$ is computed over all $n = 60$ conversations with proportional reweighting for missing D4 ($n = 57$) and D5 ($n = 58$) values. The v2.0\_current scheme $\rho = 0.291$ in Table~\ref{tab:weights} uses complete-case computation, yielding a slightly different value. Both are reported transparently; the difference reflects missing-data handling, not inconsistency.}

\paragraph{Missing data mechanism.} D4 (Objection Handling, $n = 57$) and D5 (Contextual Memory, $n = 58$) have reduced sample sizes because some short conversations contained no user objections (rendering D4 inapplicable) or insufficient multi-turn context for memory assessment (rendering D5 inapplicable). The LLM judge was instructed to return ``N/A'' rather than force a score when a dimension's behavioral indicators were absent. Missingness is plausibly related to conversation brevity but not directly to conversion status: among the 3 missing D4 cases, 1 was converted and 2 were not; among the 2 missing D5 cases, both were unconverted. We treat these as missing-at-random for composite computation (proportional reweighting of remaining dimensions). A sensitivity analysis treating missing D5 scores as the median value (3.0) yields composite $\rho = 0.268$, negligibly different from 0.272.

\paragraph{The composite is weakly positive---but the structural problem persists.} Unlike Phase~1, where the composite was negatively associated with conversion (confounded by agent type), Phase~2 shows a weak positive association ($\rho = 0.272$, $p = 0.036$, $d = +0.56$, computed with proportional reweighting for missing values). Converted conversations scored higher on average (2.82 vs.\ 2.46). This indicates that the Phase~1 ``paradox'' was substantially driven by the human-AI confound rather than a fundamental failure of quality scoring.

\paragraph{Dimension-level heterogeneity is confirmed (consistent with H1 and H2).} Two dimensions survive Bonferroni correction: D1 Need Elicitation ($\rho = 0.368$, $p_{\text{bonf}} = 0.027$, $d = +0.74$) and D3 Pacing Strategy ($\rho = 0.354$, $p_{\text{bonf}} = 0.039$, $d = +0.77$). At the other extreme, D5 Contextual Memory shows effectively zero association ($\rho = 0.018$, $p = 0.895$, $d \approx 0$). The remaining four dimensions show small positive but non-significant effects ($d = 0.27$ to $0.36$).

\paragraph{The composite underperforms its best components.} The composite ($\rho = 0.272$) is numerically weaker than either D1 ($\rho = 0.368$) or D3 ($\rho = 0.354$) individually. D5's null contribution dilutes the composite's criterion validity---exactly the structural risk that proxy metric theory predicts. We note that the difference $\Delta\rho = 0.096$ (D1 vs.\ composite) has not been tested for statistical significance; with $n = 60$, the confidence intervals for individual $\rho$ values are wide (approximately $\pm 0.20$), and the two correlations are not independent (both computed on the same sample). The composite dilution effect is therefore presented as a numerically observed pattern consistent with theory, not as a statistically confirmed difference. Bootstrap confidence intervals for $\Delta\rho$ would strengthen this claim and are a priority for future work.

\begin{figure}[t]
  \centering
  \includegraphics[width=\columnwidth]{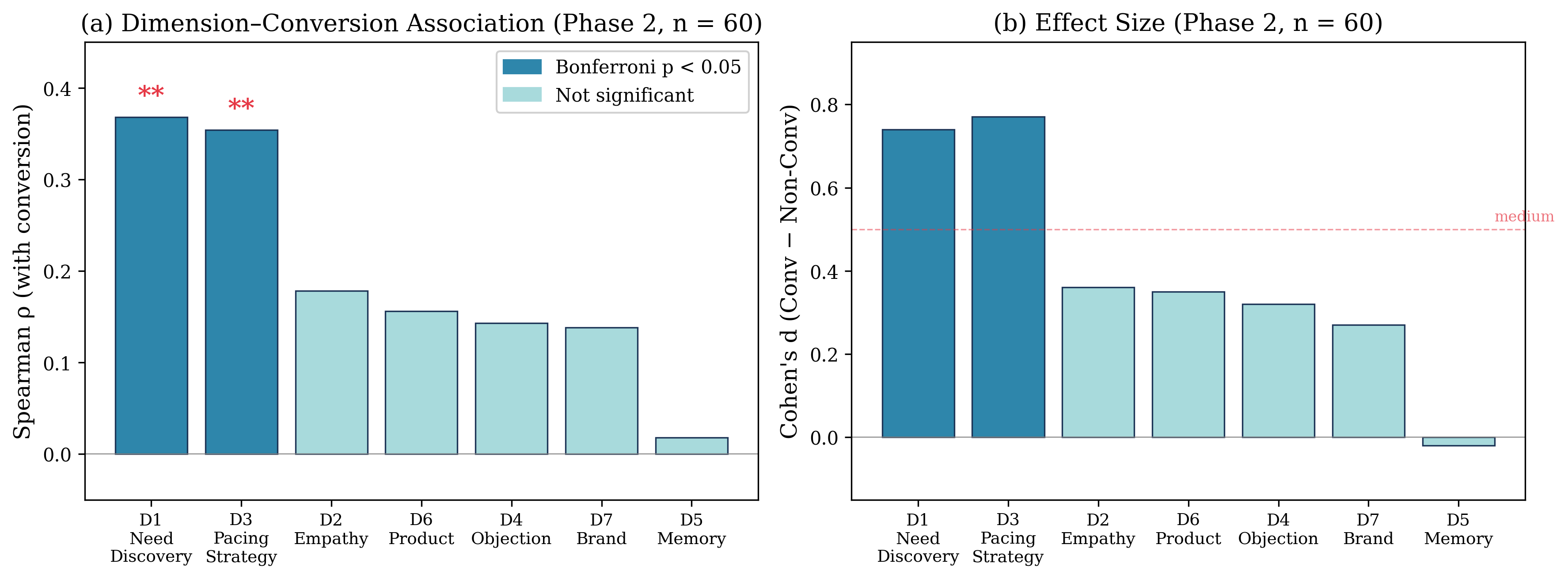}
  \caption{Phase~2 dimension--conversion analysis ($n = 60$). (a)~Spearman $\rho$: D1 and D3 survive Bonferroni correction (**). D5 shows no detectable association. (b)~Cohen's $d$: D1 and D3 show medium-to-large effects; D5 is approximately zero. Dashed line indicates medium effect threshold.}
  \label{fig:correlation}
\end{figure}

\subsection{Cross-Phase Comparison: Resolving the Paradox}
\label{sec:paradox}

Table~\ref{tab:crossphase} compares Phase~1 and Phase~2 findings, revealing how the agent-type confound shaped the pilot results.

\begin{table}[t]
  \centering
  \small
  \caption{Cross-phase comparison of key findings.}
  \label{tab:crossphase}
  \begin{tabular}{lcc}
    \toprule
    \textbf{Finding} & \textbf{Phase~1 ($n = 14$)} & \textbf{Phase~2 ($n = 60$)} \\
    \midrule
    Composite direction & No-Deal $>$ Deal \ding{55} & Conv $>$ Non-Conv \checkmark \\
    Composite $\rho$ & $+0.355$ (n.s.) & $+0.272$ ($p = 0.036$) \\
    D3 significant? & $\rho = 0.679$, $p_{\text{bonf}} = 0.054$ & $\rho = 0.354$, $p_{\text{bonf}} = 0.039^{*}$ \\
    D1 significant? & $\rho = 0.146$, n.s. & $\rho = 0.368$, $p_{\text{bonf}} = 0.027^{*}$ \\
    D5 association & $\rho = -0.284$, n.s. & $\rho = 0.018$, n.s. \\
    Agent-type confound & Present (all deals = human) & Eliminated (human only) \\
    \bottomrule
  \end{tabular}
\end{table}

The apparent ``evaluation-outcome paradox'' from Phase~1---where higher quality scores were associated with \emph{worse} outcomes---was substantially driven by the human-AI confound: AI conversations scored higher on D5 (Memory) but achieved zero conversions, dragging the composite-conversion association negative. Once this confound is removed in Phase~2, the composite reverses direction.

However, the \emph{structural} finding---dimension-level heterogeneity causing composite score dilution---is robustly confirmed. D3 remains significantly associated with conversion across both phases. D1 emerges as a second significant correlate with adequate power. D5 is confirmed as non-associated ($\rho \approx 0$). The core thesis holds: equal-weighted composites are suboptimal because they dilute conversion-associated dimensions with non-associated ones.

\subsection{What Makes D3 Distinctive}
\label{sec:d3}

Why are D1 (Need Elicitation) and D3 (Pacing Strategy) associated with conversion while D5 (Contextual Memory) is not? We offer a theoretical interpretation grounded in trust calibration theory and behavioral evidence from a complementary analysis of AI conversations.

\paragraph{Calibration dimensions vs.\ capability dimensions.} D1 and D3 both require the agent to calibrate behavior to the user's current state: D1 measures whether the agent understands user needs before acting; D3 measures whether the agent matches its pace to user readiness. D5, by contrast, measures a technical capability (information retention) that does not inherently require attunement to the user. In Lee \& See's~\citep{lee2004} trust framework, D1 and D3 demonstrate trustworthy \emph{purpose} (the agent acts in the user's interest), while D5 demonstrates \emph{performance} (the agent is technically capable). Our data suggest that purpose-related dimensions are more strongly associated with conversion in trust-dependent interactions---consistent with Hancock et al.'s~\citep{hancock2011} finding that robot performance characteristics explain more trust variance than human or environmental factors.

\paragraph{Behavioral evidence from AI conversations.} Complementary qualitative analysis of 10 AI conversations (Phase~1 data) illustrates the extreme case of low D3: in 6/10, the user rejected the sales pitch $\geq 3$ times while the AI continued pushing; AI conversations averaged 89.57 funnel stage transitions vs.\ human conversations' 26.23, indicating cyclical repetition rather than linear progression; in 7/10, the agent expressed empathy immediately followed by a purchase link. These patterns represent a failure of \emph{calibration}---the agent executes correct behaviors at incorrect times.

\paragraph{Confound control.} D3's association with conversion could be partially driven by conversation length (longer conversations mechanically lower D3 scores due to more rejection opportunities) or user initial intent. We directly test the conversation-length confound through logistic regression in Section~\ref{sec:robustness}: D3's odds ratio \emph{increases} from 2.71 to 3.18 after controlling for message count ($p = 0.006$), indicating that the association is not an artifact of conversation length. Notably, Phase~2 also eliminates the agent-type confound that was a serious concern in Phase~1 (where all deal cases were human). User initial intent remains uncontrolled.

\subsection{Robustness: Multivariate Analysis and Confound Control}
\label{sec:robustness}

A key concern is whether the observed dimension--conversion associations survive after controlling for confounding variables, particularly conversation length (message count), which correlates with both quality scores and conversion opportunity. We address this through logistic regression and partial correlation analysis.

\paragraph{Logistic regression with conversation length control.} Table~\ref{tab:logistic} presents logistic regression results for D1 and D3 (the two Bonferroni-significant dimensions) and D5 (the null dimension), both univariate and controlling for message count.

\begin{table}[t]
  \centering
  \small
  \caption{Logistic regression: dimension--conversion association controlling for conversation length ($n = 60$).}
  \label{tab:logistic}
  \begin{tabular}{lcccc}
    \toprule
    \textbf{Model} & \textbf{OR} & \textbf{95\% CI} & \textbf{$p$} & \textbf{AIC} \\
    \midrule
    D3 (univariate) & 2.71 & [1.28, 5.75] & 0.009 & 77.2 \\
    D3 + msg\_count & \textbf{3.18} & \textbf{[1.39, 7.24]} & \textbf{0.006} & 71.9 \\
    \midrule
    D1 (univariate) & 2.58 & [1.24, 5.36] & 0.011 & 77.9 \\
    D1 + msg\_count & 2.49 & [1.17, 5.30] & 0.018 & 75.2 \\
    \midrule
    D5 (univariate) & 0.97 & [0.55, 1.74] & 0.931 & 83.3 \\
    \midrule
    Full (D1+D3+D5+msg) & --- & --- & --- & \textbf{60.7} \\
    \quad D1 & 7.06 & [1.89, 26.33] & 0.004 & \\
    \quad D3 & 5.66 & [1.68, 19.05] & 0.005 & \\
    \quad D5 & 0.15 & [0.04, 0.56] & 0.005 & \\
    \bottomrule
  \end{tabular}
\end{table}

Three findings strengthen the criterion validity evidence. First, D3's odds ratio \emph{increases} from 2.71 (univariate) to 3.18 (controlling for message count, $p = 0.006$): each one-point increase in D3 is associated with a 3.18$\times$ increase in the odds of conversion, after accounting for conversation length. The effect strengthens rather than attenuates, indicating that the D3--conversion association is not an artifact of longer conversations. Second, D1 survives confound control (OR = 2.49, $p = 0.018$). Third, D5 remains non-significant (OR = 0.97, $p = 0.931$), confirming its null association. All variance inflation factors are below 2.0, indicating no problematic multicollinearity.

\paragraph{Partial Spearman correlations.} Controlling for message count, D3's partial correlation with conversion remains significant ($r = 0.355$, $p = 0.006$, df $= 57$) and D1's partial correlation likewise persists ($r = 0.332$, $p = 0.010$). The near-identical magnitude to the zero-order correlations ($\rho = 0.354$ and $0.368$) indicates that conversation length does not confound the dimension--conversion relationships.

\paragraph{Interpretation.} The full multivariate model (D1 + D3 + D5 + message count; AIC = 60.7) substantially outperforms any single-predictor model, suggesting that D1 and D3 capture complementary aspects of conversation quality relevant to conversion. Notably, D5 becomes a significant \emph{negative} predictor in the multivariate model (OR = 0.15, $p = 0.005$), suggesting that after controlling for D1 and D3, high contextual memory scores may be associated with lower conversion---consistent with the hypothesis that technical capability without calibration is at best irrelevant and potentially counterproductive. We note, however, that this suppressor effect should be interpreted cautiously: the full model has an events-per-variable ratio of 5.0 (25 events / 5 parameters), at the lower bound of the 10:1 guideline for stable logistic regression~\citep{peduzzi1996}, and the wide confidence intervals (D1: [1.89, 26.33]; D3: [1.68, 19.05]) reflect parameter instability. The bivariate models (one predictor + message count; EPV $\geq 8$) provide more reliable estimates.

\section{Restoring Criterion Validity: Three-Layer Architecture}
\label{sec:architecture}

\subsection{Architecture Design}
\label{sec:archdesign}

\begin{figure}[t]
  \centering
  \includegraphics[width=0.70\columnwidth]{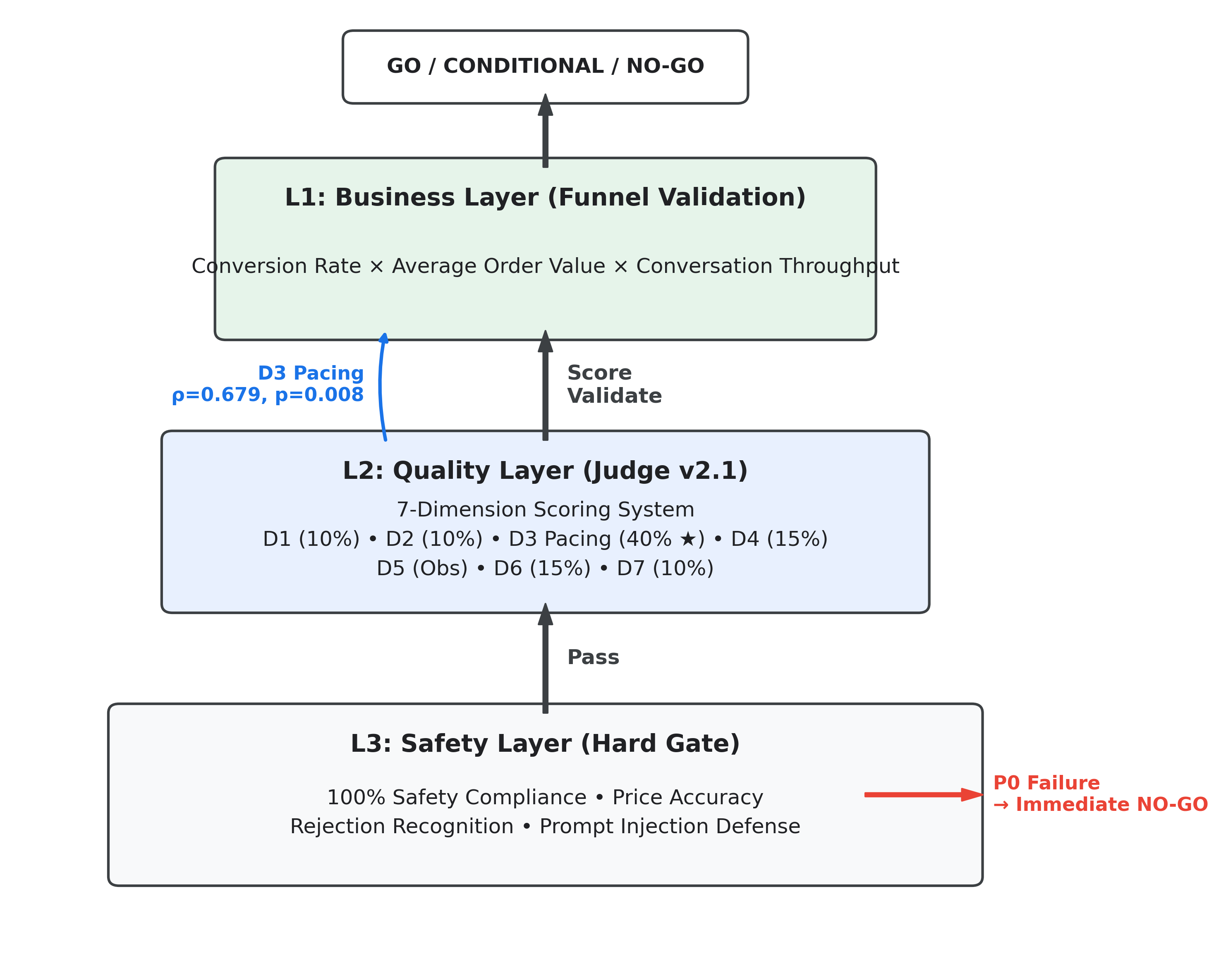}
  \caption{Three-layer evaluation architecture. L3~Safety (hard gate) $\to$ L2~Quality (LLM-as-Judge with conversion-informed weights) $\to$ L1~Task Success (criterion layer). D1 and D3 provide the empirically observed bridge between L2 and L1 ($\rho = 0.368$ and $\rho = 0.354$ respectively, Phase~2).}
  \label{fig:arch}
\end{figure}

Based on the criterion validity analysis, we propose a three-layer evaluation architecture that explicitly separates safety, quality, and business concerns (Figure~\ref{fig:arch}).

\paragraph{L3 (Safety Layer).} A hard gate covering safety compliance, price accuracy, rejection recognition, and prompt injection defense. Any P0 failure results in immediate NO-GO.

\paragraph{L2 (Quality Layer).} LLM-as-Judge scoring with conversion-informed weights. The key change from standard practice is weighting dimensions by their empirically observed association with the outcome criterion, rather than by expert intuition or equal weighting.

\paragraph{L1 (Business Layer).} Direct measurement of conversion funnel metrics. L1 serves as the ultimate criterion against which L2's validity is assessed: if L2 improvements do not translate to L1 improvements, the evaluation framework needs recalibration.

\subsection{Conversion-Informed Weight Exploration (H3 Examined In-Sample)}
\label{sec:weights}

We tested 6 weight schemes against conversion labels across both study phases (Table~\ref{tab:weights}).

\textbf{Critical caveat on circular analysis}: In Phase~1, the dimension-conversion correlations that motivated the weight redesign and the weight optimization itself were computed on the same $n = 14$ dataset---a well-known form of circular analysis~\citep{kriegeskorte2009}. Phase~2 provides a partial remedy: the weight schemes designed from Phase~1 data are now tested on independently sampled Phase~2 data ($n = 60$). However, some circularity remains because we also report new Phase~2 scheme comparisons on the same Phase~2 dataset. We partially address this through 4-fold temporal cross-validation (Section~\ref{sec:circular}), but prospective validation on fully independent data is still needed.

\begin{table}[t]
  \centering
  \small
  \caption{Weight scheme comparison: Phase~1 ($n = 14$) and Phase~2 ($n = 60$).}
  \label{tab:weights}
  \begin{tabular}{lccccccccccc}
    \toprule
     & & & & & & & & \multicolumn{2}{c}{\textbf{Phase 1}} & \multicolumn{2}{c}{\textbf{Phase 2}} \\
    \cmidrule(lr){9-10}\cmidrule(lr){11-12}
    \textbf{Scheme} & D1 & D2 & D3 & D4 & D5 & D6 & D7 & $\rho$ & $p$ & $\rho$ & $p$ \\
    \midrule
    \textbf{conversion\_informed} & 10 & 10 & \textbf{40} & 15 & \textbf{0} & 15 & 10 & 0.607 & 0.021 & \textbf{0.351} & \textbf{0.006**} \\
    d3\_boosted\_40 & 10 & 10 & 40 & 15 & 10 & 10 & 5 & 0.570 & 0.033 & 0.329 & 0.010* \\
    d3\_boosted\_30 & 10 & 15 & 30 & 15 & 10 & 10 & 10 & 0.478 & 0.084 & 0.304 & 0.018* \\
    d5\_removed & 15 & 15 & 20 & 15 & 0 & 20 & 15 & 0.465 & 0.094 & 0.302 & 0.019* \\
    v2.0\_current & 20 & 20 & 20 & 15 & 10 & 10 & 5 & 0.355 & 0.213 & 0.291 & 0.024* \\
    v1\_equal & 14 & 14 & 14 & 14 & 14 & 14 & 14 & 0.282 & 0.329 & 0.230 & 0.078 \\
    \bottomrule
  \end{tabular}
\end{table}

Phase~2 results strengthen the case for conversion-informed weighting. The conversion\_informed scheme achieves the highest criterion validity ($\rho = 0.351$, $p = 0.006$), substantially outperforming equal weighting ($\rho = 0.230$, $p = 0.078$, n.s.). The rank ordering of schemes is largely preserved across phases, suggesting that the Phase~1 signal was not purely noise. Two operations continue to drive improvement: boosting D3 from 20\% to 40\% and removing D5 (setting its weight to 0\%)---consistent with the dimension-level findings in Section~\ref{sec:h2}.

\begin{figure}[t]
  \centering
  \includegraphics[width=0.85\columnwidth]{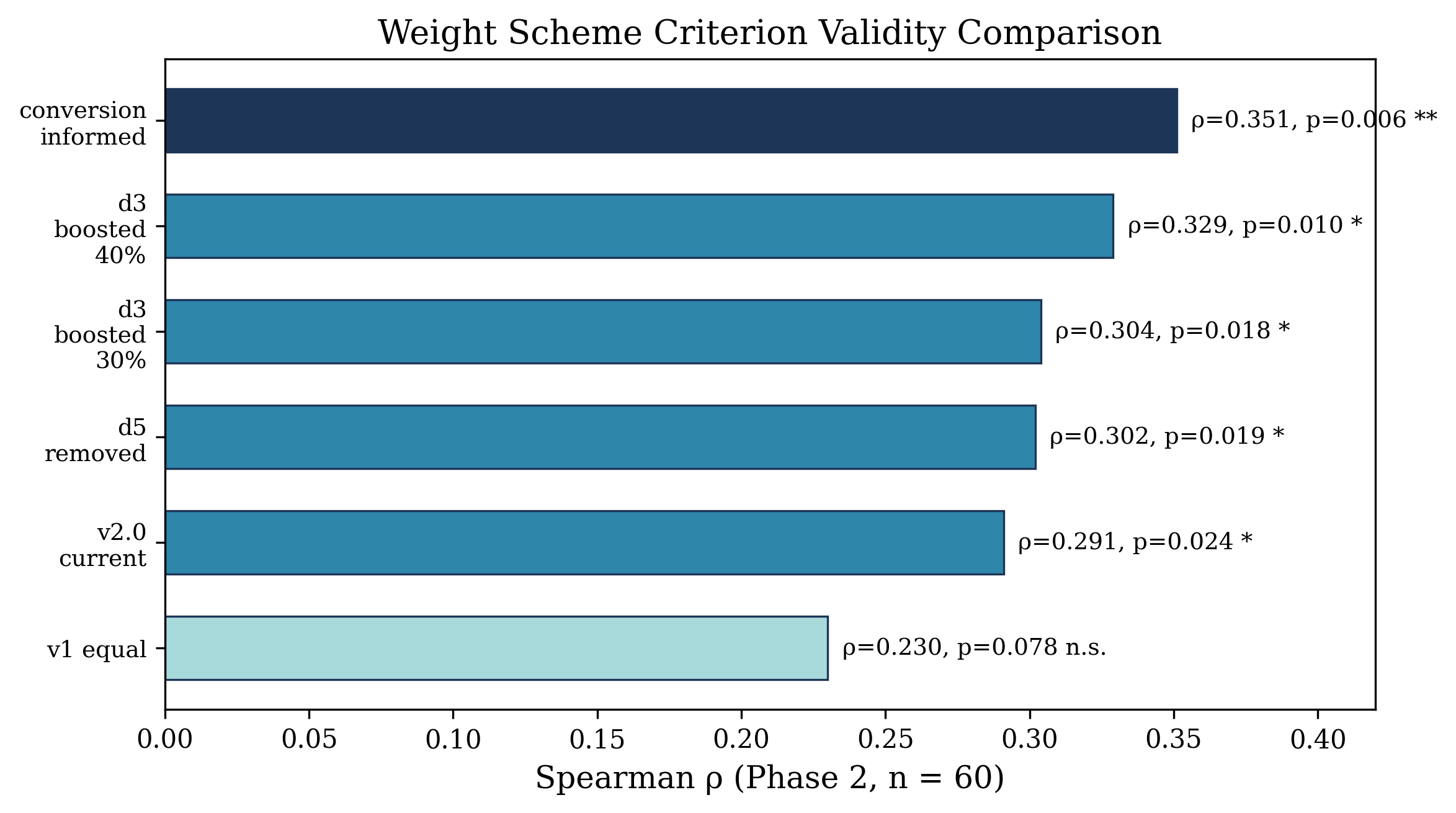}
  \caption{Weight scheme criterion validity comparison (Phase~2, $n = 60$). All schemes except equal weighting reach $p < 0.05$; the conversion-informed scheme achieves the highest $\rho = 0.351$ ($p = 0.006$).}
  \label{fig:weights}
\end{figure}

\subsection{In-Sample Score Direction Reversal}
\label{sec:restoration}

Table~\ref{tab:restoration} compares group means under different weighting schemes in Phase~2.

\begin{table}[t]
  \centering
  \small
  \caption{Phase~2: Group means under v2.0 (equal-ish) vs.\ conversion-informed weights.}
  \label{tab:restoration}
  \begin{tabular}{lccc}
    \toprule
    \textbf{Group} & \textbf{v2.0 Score} & \textbf{Conv.-Informed} & \textbf{$\Delta$} \\
    \midrule
    Converted ($n = 25$) & 2.82 & 2.80 & \\
    Not-Converted ($n = 35$) & 2.46 & 2.35 & \\
    \midrule
    Group gap & $+0.36$ \checkmark & $\mathbf{+0.45}$ \checkmark & $+0.09$ \\
    Spearman $\rho$ & 0.272 ($p = 0.036$) & \textbf{0.351} ($p = 0.006$) & $+0.079$ \\
    \bottomrule
  \end{tabular}
\end{table}

Unlike Phase~1---where the composite direction was \emph{reversed} (No-Deal $>$ Deal) and conversion-informed weighting was needed to flip the sign---Phase~2 shows the correct direction under \emph{all} weighting schemes. Conversion-informed weighting now serves a different but still important function: it \emph{amplifies} an existing positive signal rather than correcting a negative one, widening the group gap from 0.36 to 0.45 and improving $\rho$ from 0.272 to 0.351.

\begin{figure}[t]
  \centering
  \includegraphics[width=\columnwidth]{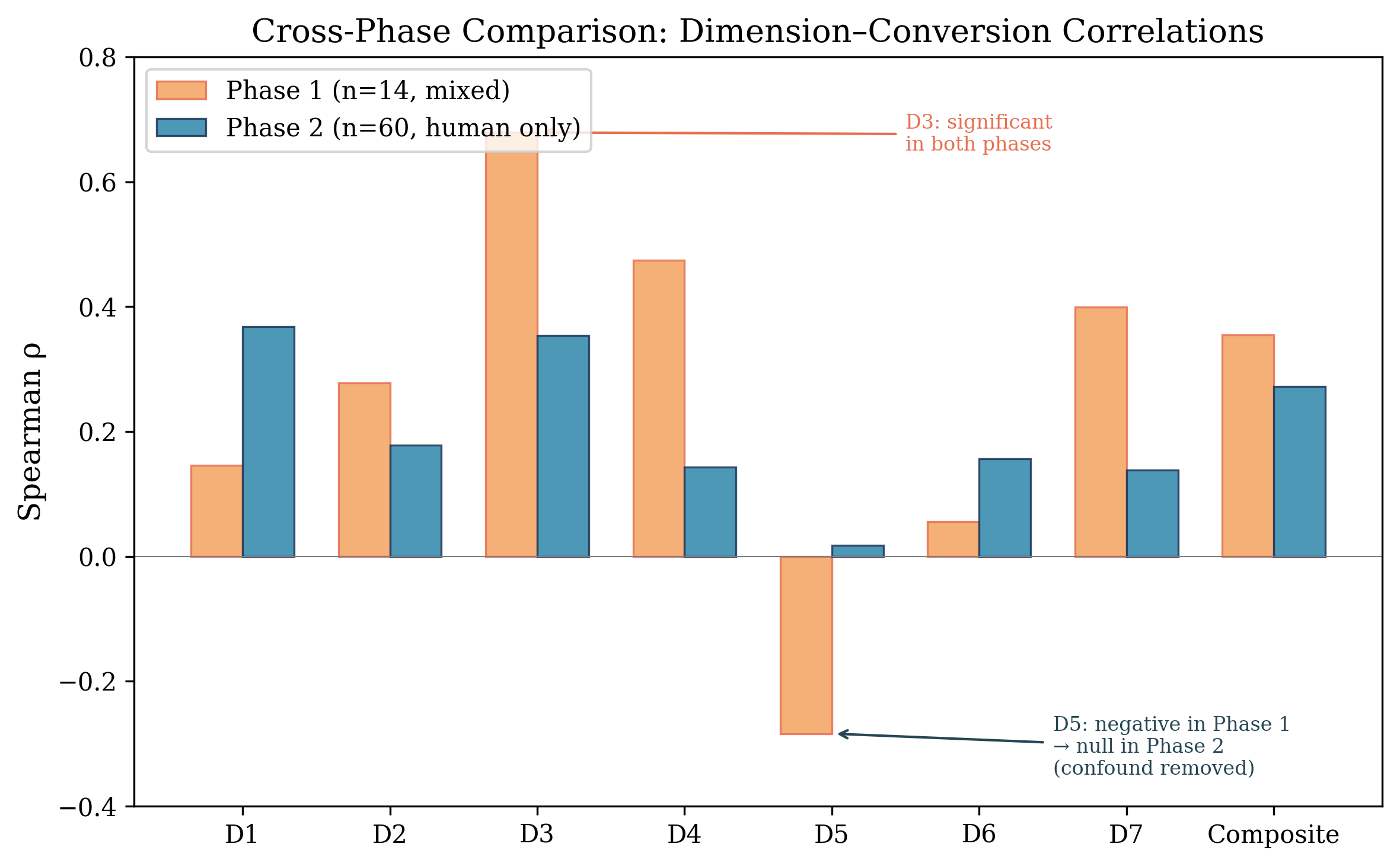}
  \caption{Cross-phase comparison of dimension--conversion correlations. D3 is significant in both phases. D5 shifts from negative (Phase~1, confounded) to null (Phase~2, confound removed). D1 emerges as a second significant correlate with adequate power in Phase~2.}
  \label{fig:paradox}
\end{figure}

\subsection{D3 Hard-Cap Rule}
\label{sec:hardcap}

To enforce pacing discipline, v2.1 introduces hard-cap rules (Table~\ref{tab:caps}).

\begin{table}[t]
  \centering
  \small
  \caption{D3 hard-cap rules.}
  \label{tab:caps}
  \begin{tabular}{lcc}
    \toprule
    \textbf{Violation} & \textbf{D3 Cap} & \textbf{Total Cap} \\
    \midrule
    Same message 3+ consecutive days & 2 & --- \\
    User rejects $\geq$3 times, agent continues & \textbf{1} & \textbf{2.4} \\
    User rejects $\geq$5 times, agent continues & \textbf{1} & \textbf{2.0} \\
    Purchase link on every message & 2 & --- \\
    \bottomrule
  \end{tabular}
\end{table}

These caps operationalize an ethical principle: persistence beyond clear rejection is not selling---it is harassment.

\section{The Trust-Funnel Framework: Behavioral Mechanism}
\label{sec:trustfunnel}

\subsection{Behavioral Mechanism: The Behavior-Trust Desynchronization}

The criterion validity analysis (Section~\ref{sec:validity}) establishes that quality dimensions differ in their association with conversion, with calibration dimensions (D1, D3) showing significant associations while capability dimensions (D5) do not. The Trust-Funnel framework provides a \emph{candidate mechanistic hypothesis} for understanding \emph{why}: sales behavior and user trust operate on parallel but distinct tracks that can become desynchronized, particularly for AI agents.

\emph{Methodological caveat:} The evidence in this section is descriptive and observational, drawn from non-equivalent samples (100 AI conversations from PolarDB vs.\ 30 pre-selected ``golden'' human conversations). The AI and human groups differ in sampling logic, conversation context, and selection criteria, so quantitative comparisons (e.g., stage transition counts) should be interpreted as illustrative patterns rather than controlled comparisons. The Trust Ladder annotations are LLM-generated without human validation (Section~\ref{sec:trustval}). We present this framework as a plausible mechanism warranting further investigation, not as a statistically established causal pathway.

The \textbf{Funnel Track} (F1--F6) captures what the agent \emph{does}; the \textbf{Trust Track} (T0--T5) captures what the user \emph{experiences}. The behavioral analysis (130 conversations: 100 AI + 30 human) suggests that these tracks are dramatically desynchronized for the AI agent (Table~\ref{tab:desynch}).

\begin{table}[t]
  \centering
  \small
  \caption{Behavior-trust desynchronization metrics.}
  \label{tab:desynch}
  \begin{tabular}{lccc}
    \toprule
    \textbf{Metric} & \textbf{AI ($n=100$)} & \textbf{Human ($n=30$)} & \textbf{Ratio} \\
    \midrule
    Mean stage transitions & \textbf{89.57} & 26.23 & \textbf{3.42$\times$} \\
    F6 (Closing) reach rate & 72\% & 97\% & 0.74$\times$ \\
    Messages at F6 stage & 25.1 & 4.5 & 5.6$\times$ \\
    \bottomrule
  \end{tabular}
\end{table}

The AI executes the sales funnel with high coverage (72\% reach closing) but fails on the trust track (0\% reach T5 in the scored sample). In Lee \& See's~\citep{lee2004} framework, this represents trust \emph{miscalibration}.

\begin{figure}[t]
  \centering
  \includegraphics[width=0.65\columnwidth]{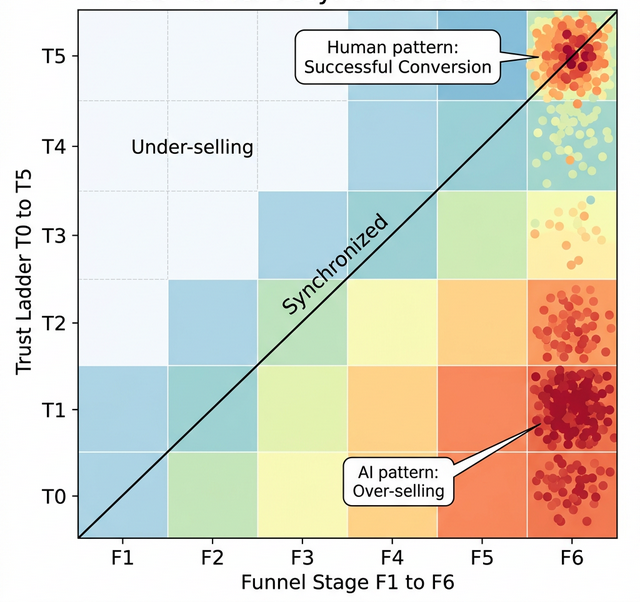}
  \caption{Trust-Funnel desynchronization matrix. Human conversations cluster in upper-right (F6 $\times$ T5---synchronized). AI conversations cluster in lower-right (F6 $\times$ T0--T2---desynchronized: selling without trust).}
  \label{fig:trustfunnel}
\end{figure}

\subsection{Trust Ladder Patterns from Human Agents}

Annotation of 30 human-agent conversations revealed five conversion patterns (Table~\ref{tab:patterns}).

\begin{table}[t]
  \centering
  \small
  \caption{Trust Ladder conversion patterns in human agent dialogues ($n = 30$).}
  \label{tab:patterns}
  \begin{tabular}{lcp{5.5cm}}
    \toprule
    \textbf{Pattern} & \textbf{Freq.} & \textbf{Key Characteristic} \\
    \midrule
    Repeated Hesitation & 37\% & User oscillates at T4/T5, requires reassurances \\
    Quick Decision & 23\% & Skips stages, driven by pre-existing need \\
    Standard Cultivation & 20\% & Textbook T1$\to$T2$\to$T3$\to$T4$\to$T5 \\
    Price-Sensitive & 13\% & Primary bottleneck at T5, price negotiation \\
    Trust-Driven & 7\% & Bottleneck at T2/T3, rapid post-trust conversion \\
    \bottomrule
  \end{tabular}
\end{table}

\begin{figure}[t]
  \centering
  \includegraphics[width=0.75\columnwidth]{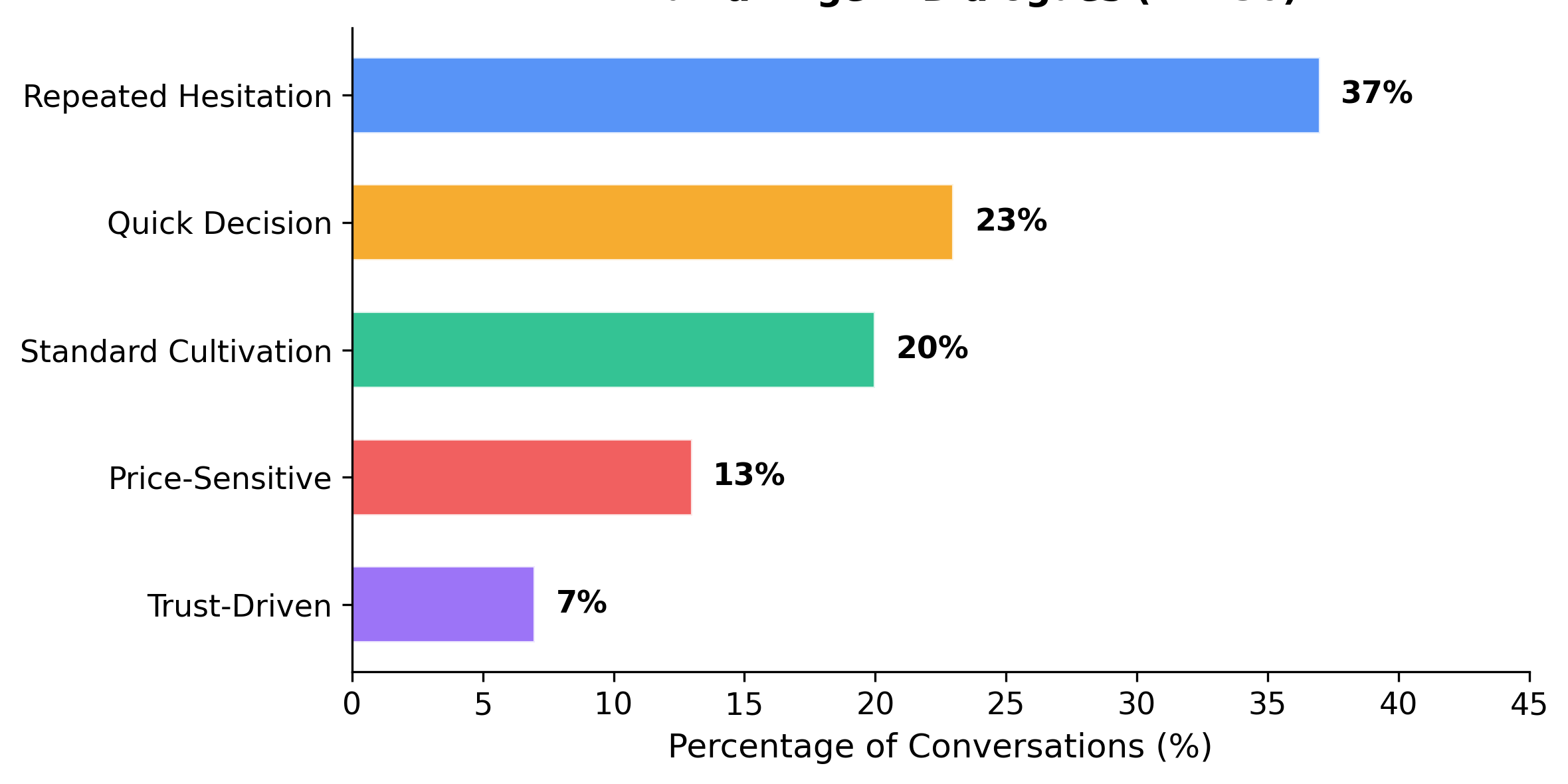}
  \caption{Trust Ladder conversion patterns in human agent dialogues ($n = 30$).}
  \label{fig:patterns}
\end{figure}

\subsection{The Trust Gate: A Design Principle}

Based on the desynchronization finding, we propose the \textbf{Trust Gate}---a mechanism constraining funnel progression based on verified trust signals (Table~\ref{tab:trustgate}).

\begin{table}[t]
  \centering
  \small
  \caption{Trust Gate: permissible funnel stages by verified trust state.}
  \label{tab:trustgate}
  \begin{tabular}{ll}
    \toprule
    \textbf{Trust State Verified} & \textbf{Permitted Stages} \\
    \midrule
    T1 (Platform Credible) & F1, F2 only \\
    T2 (Agent Helpful) & F1--F4 \\
    T3 (Success Evidence) & F1--F5 \\
    T4 (Deep Consideration) & F1--F6 \\
    Below T1 & F1 only \\
    \bottomrule
  \end{tabular}
\end{table}

This mechanism addresses the AI's dominant failure mode: in 50\% of AI conversations reaching F6, the user was still at T1.

\section{Failure-Driven Optimization Flywheel}
\label{sec:flywheel}

\emph{Note: This section describes operational methodology derived from the criterion validity findings. The empirical evidence here is limited to two improvement cycles (case-level, not statistically validated). We include it to illustrate the full path from diagnosis to deployment, but the primary empirical contribution of this paper is the criterion validity analysis in Sections~\ref{sec:validity}--\ref{sec:architecture}.}

\subsection{From Diagnosis to Improvement}

\begin{figure}[t]
  \centering
  \includegraphics[width=0.60\columnwidth]{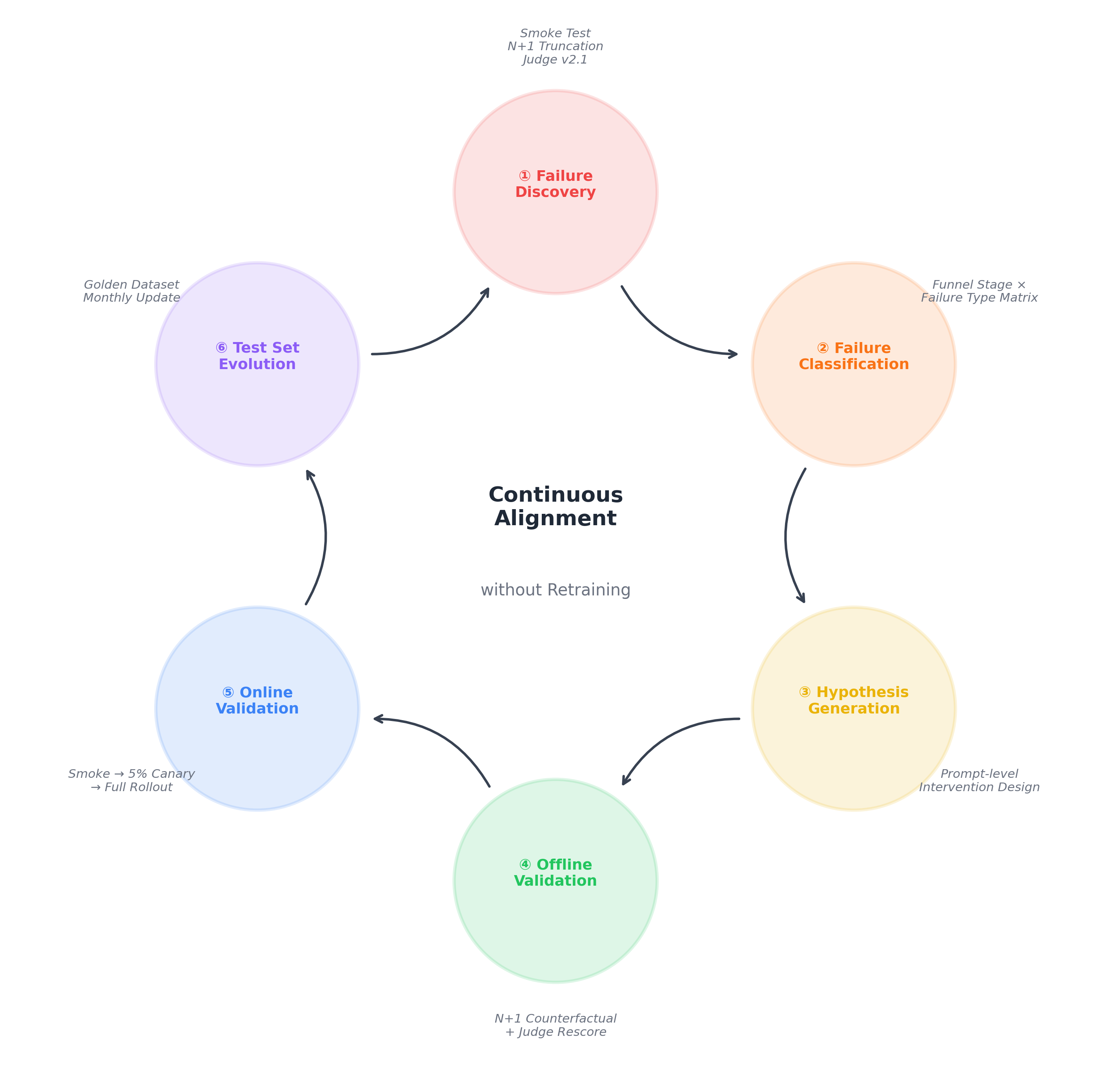}
  \caption{Failure-Driven Optimization Flywheel: six-stage closed-loop for continuous alignment without retraining.}
  \label{fig:flywheel}
\end{figure}

The criterion validity analysis and Trust-Funnel framework provide diagnostic power. This section describes a \emph{proposed} methodology for translating diagnosis into systematic improvement via a six-stage closed-loop cycle (Figure~\ref{fig:flywheel}): Failure Discovery $\to$ Failure Classification $\to$ Hypothesis Generation $\to$ Offline Validation $\to$ Online Validation $\to$ Test Set Evolution.

\subsection{Multi-Granularity Failure Discovery}

Failures are surfaced at three granularities (Table~\ref{tab:granularity}).

\begin{table}[t]
  \centering
  \small
  \caption{Multi-granularity failure discovery mechanisms.}
  \label{tab:granularity}
  \begin{tabular}{llcl}
    \toprule
    \textbf{Mechanism} & \textbf{Granularity} & \textbf{Speed} & \textbf{Catches} \\
    \midrule
    Smoke Test (47 cases) & Scenario & 5--10 min & Known patterns \\
    N+1 Truncation (50 pts) & Turn & ${\sim}$1 hour & Single-response quality \\
    Judge v2.1 (7 dims) & Conversation & ${\sim}$2 hours & Strategic failures \\
    \bottomrule
  \end{tabular}
\end{table}

The N+1 evaluation isolates \emph{response quality} from \emph{strategic quality}. At 50 truncation points from 21 golden conversations, the AI achieved near-parity on single turns (19 wins vs.\ 19 losses, 12 ties) while failing on multi-turn strategy---confirming that the problem is \emph{when} and \emph{how often} the AI acts, not \emph{what} it says at any given moment. Notably, the AI scored \emph{higher} on empathy (R2: 3.86 vs.\ 2.88) and relevance (R1: 4.48 vs.\ 4.00) but \emph{lower} on naturalness (R4: 3.68 vs.\ 4.26), suggesting that single-turn quality metrics mask multi-turn strategic failures---the same composite dilution effect observed at the conversation level.

\begin{figure}[t]
  \centering
  \includegraphics[width=\columnwidth]{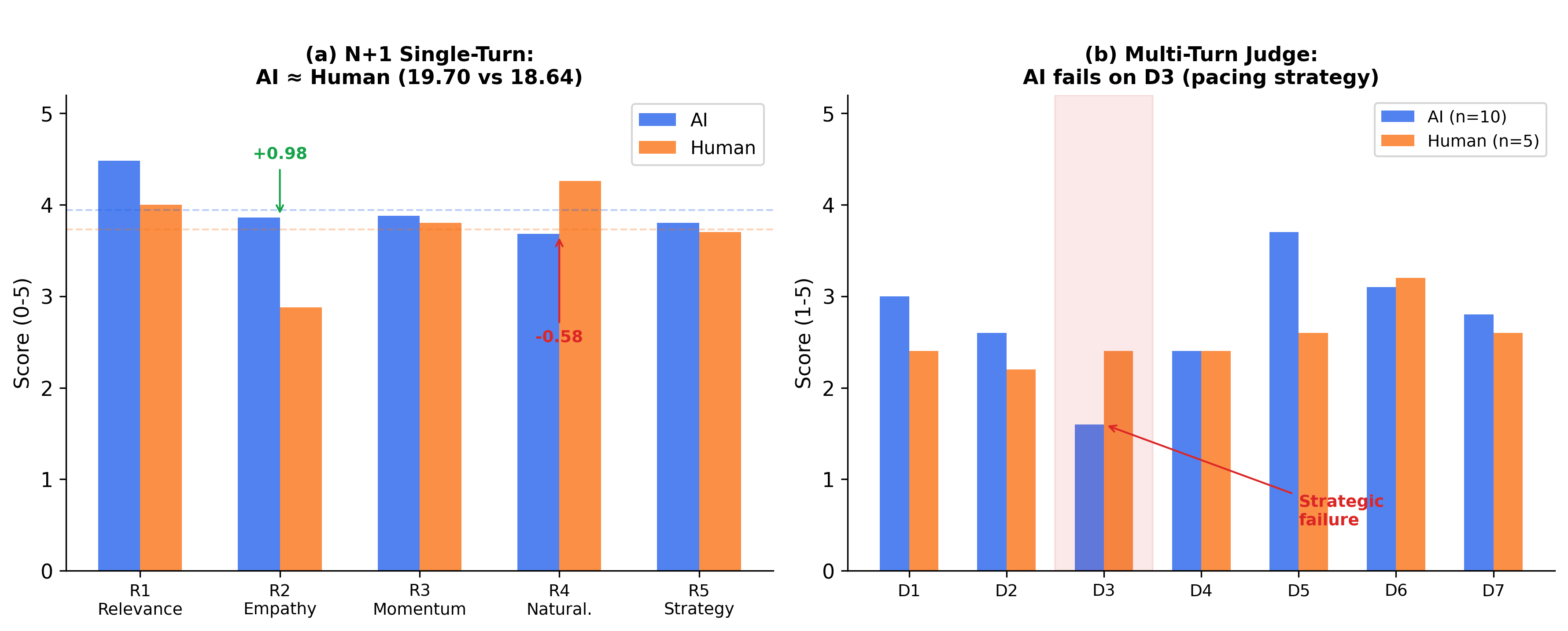}
  \caption{The single-turn/multi-turn performance gap. (a)~N+1 single-turn: AI $\approx$ Human. (b)~Multi-turn Judge: AI fails on D3 (pacing strategy).}
  \label{fig:singlemulti}
\end{figure}

\subsection{Failure Classification Matrix}

\begin{figure}[t]
  \centering
  \includegraphics[width=0.70\columnwidth]{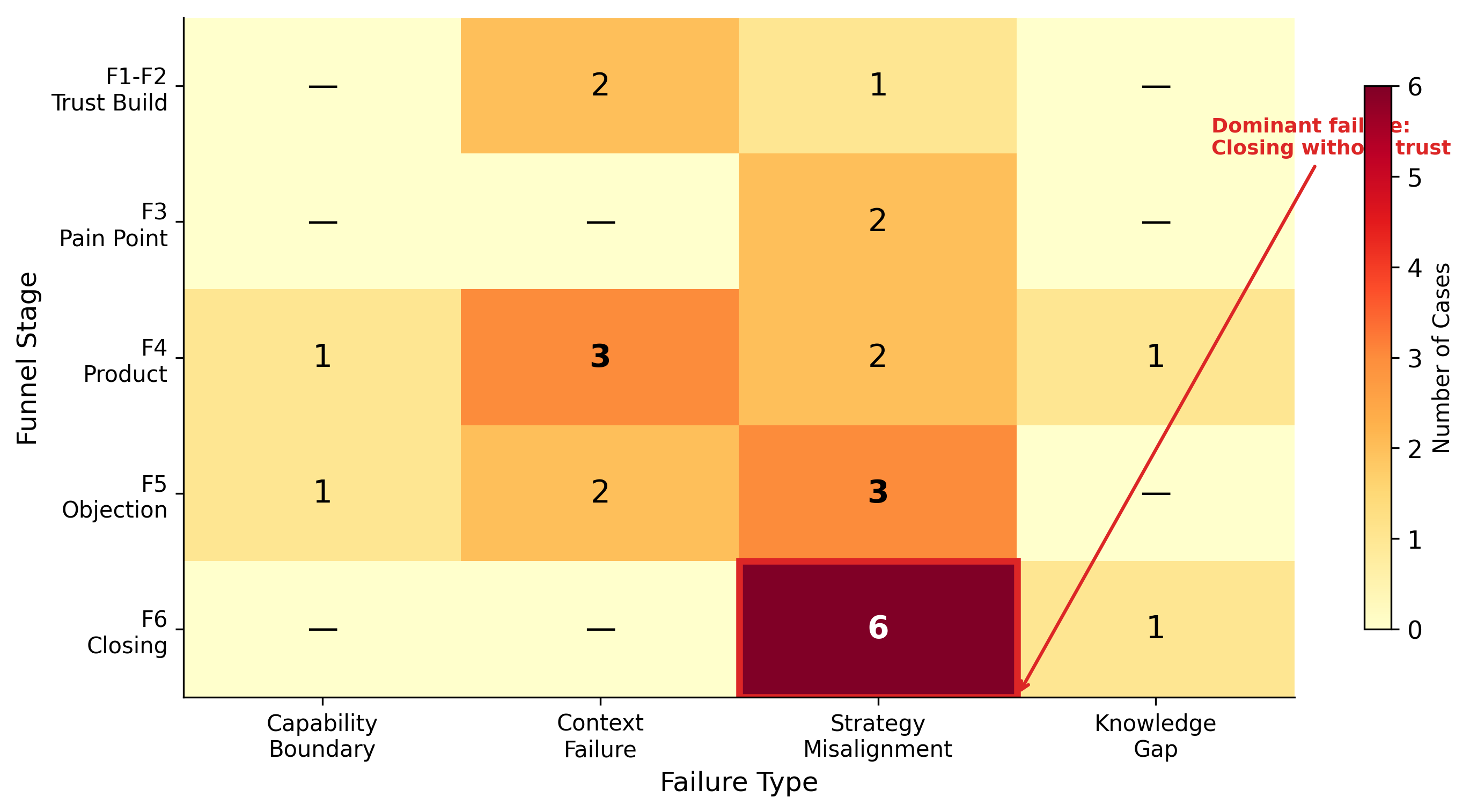}
  \caption{Funnel $\times$ failure type matrix. Dominant cell: F6 (Closing) $\times$ Strategy Misalignment = 6/10 cases.}
  \label{fig:matrix}
\end{figure}

Failures are classified along two axes: funnel stage (where) $\times$ failure type (what) (Figure~\ref{fig:matrix}).

\subsection{Prompt-Level Intervention Design}

Each classified failure generates a testable hypothesis and prompt-level intervention (Table~\ref{tab:interventions}).

\begin{table}[t]
  \centering
  \small
  \caption{Prompt-level interventions derived from failure classification.}
  \label{tab:interventions}
  \begin{tabular}{lll}
    \toprule
    \textbf{Failure Pattern} & \textbf{Intervention} & \textbf{Mechanism} \\
    \midrule
    F6 without T4+ trust & Trust Gate rules & Constrain closing \\
    Rejection ignored $\geq$3$\times$ & 3-tier classification & Soft$\to$hard$\to$terminal \\
    F4/F5 oscillation & Per-stage msg caps & Anti-cycling rules \\
    Empathy$\to$sales trap & Temporal decoupling & 3-msg buffer \\
    \bottomrule
  \end{tabular}
\end{table}

\subsection{Relationship to RLHF/DPO}

The flywheel complements rather than replaces weight-level optimization. Prompt-level changes iterate in hours (vs.\ days for RLHF/DPO), operate at specific-rule granularity (vs.\ broad tendencies), and offer instant rollback. When the flywheel identifies stable patterns (e.g., D3 $\geq 3$ consistently outperforms), these become candidate training signals for eventual DPO fine-tuning~\citep{rafailov2023}.

\subsection{Operationalization Evidence: Two Evaluation Cycles}
\label{sec:opevidence}

To demonstrate the practical utility of the three-layer architecture, we report results from two iterative evaluation cycles comparing candidate AI prompt configurations (Table~\ref{tab:evalcycles}).

\begin{table}[t]
  \centering
  \small
  \caption{Three-layer evaluation results across two improvement cycles.}
  \label{tab:evalcycles}
  \begin{tabular}{lcccccc}
    \toprule
    & \multicolumn{3}{c}{\textbf{Cycle 1}} & \multicolumn{3}{c}{\textbf{Cycle 2}} \\
    \cmidrule(lr){2-4}\cmidrule(lr){5-7}
    & \textbf{Config A} & \textbf{Config B} & & \textbf{Config A} & \textbf{Config B} & \\
    \midrule
    L3 P0 pass rate & 88.9\% & 94.4\% & & 94.4\% & \textbf{100\%} & \\
    L3 decision & NO-GO & NO-GO & & NO-GO & \textbf{GO} \\
    \midrule
    D3 mean (v2.1) & 2.69 & 2.56 & & 2.84 & 2.70 \\
    D3 vs.\ baseline & --- & --- & & +0.15 & +0.14 \\
    Weighted total & 2.71 & 2.64 & & 2.99 & 2.91 \\
    \midrule
    Key L3 failures & S31, S33 & S37 & & S31 & none \\
    \bottomrule
  \end{tabular}
\end{table}

Cycle~1 established the baseline: both configurations failed L3 safety gates (P0 pass rate $< 100\%$, primarily on rejection recognition scenarios). Neither was cleared for deployment. Cycle~2 incorporated the P0 interventions from Section~\ref{sec:flywheel}: three-tier rejection classification (soft/hard/terminal), per-stage message caps, and Trust Gate constraints. Config~B achieved 100\% P0 pass rate and was approved for deployment (GO decision). D3 improved from baseline 2.56 to 2.70, and the weighted total from 2.64 to 2.91.

Two observations are noteworthy. First, L3 acts as a genuine hard gate: Config~A in Cycle~2 scored higher on L2 quality (D3 = 2.84 vs.\ 2.70) but was rejected because it failed one P0 safety case (soft rejection recognition). Quality improvements are insufficient without safety compliance---the architecture enforces this by design. Second, the D3 improvement from 1.60 (AI baseline, Phase~1) to 2.70 (Cycle~2 Config~B) demonstrates that the criterion validity findings translate into actionable prompt-level improvements.

\section{Discussion}
\label{sec:discussion}

\subsection{Theoretical Implications}

\paragraph{Criterion validity as standard practice.} Our two-phase study provides converging evidence that multi-dimensional quality scores, while directionally valid in aggregate, suffer from dimension-level heterogeneity that degrades composite criterion validity. The current practice---validating metrics against human preferences or inter-rater agreement---establishes construct validity and reliability but does not ensure that scores are optimally associated with the outcomes they are meant to serve. We propose that evaluation frameworks in applied settings should include an explicit \emph{outcome alignment validation step}: (1)~score conversations on quality dimensions; (2)~measure the downstream outcome; (3)~compute dimension-outcome associations; (4)~adjust weighting based on empirical associations. This procedure is not novel---it is standard in psychometric instrument development~\citep{messick1995}---but it is conspicuously absent from dialogue evaluation practice. Our Phase~2 results demonstrate that even a simple reweighting (D3 = 40\%, D5 = 0\%) can improve $\rho$ from 0.272 to 0.351 ($p = 0.006$).

\paragraph{What is being validated: rubric design vs.\ judge capability.} An important distinction: our findings primarily concern the criterion validity of the \emph{evaluation rubric} (which dimensions to measure and how to weight them), not the criterion validity of the \emph{judge method} (whether the LLM scores accurately). D5's null association with conversion is not because the LLM judge mismeasures contextual memory---it is because contextual memory, even when accurately measured, is not associated with conversion in this context. Any judge (human or LLM) using the same rubric would face the same dimension heterogeneity. This distinction matters because it redirects the intervention: the solution is not a better judge but a better-weighted rubric informed by outcome data.

\paragraph{Dimension-level heterogeneity as a general risk.} The mechanism underlying criterion validity degradation---individual dimensions having heterogeneous relationships with the outcome---is not specific to our context. In Phase~2, D1 and D3 show moderate-to-large effects ($d = 0.74$ and $0.77$), while D5 shows effectively zero effect ($d \approx 0$). Any multi-dimensional evaluation exhibiting similar heterogeneity will produce a composite score with suboptimal criterion validity. This risk exists in customer service evaluation (CSAT components vs.\ resolution), educational assessment (engagement vs.\ learning), and therapeutic dialogue (empathy scores vs.\ symptom improvement). The structural possibility of this failure mode warrants investigation across applied dialogue domains.

\paragraph{Implications for preference-optimized training.} Reinforcement Learning from Human Feedback (RLHF) and Direct Preference Optimization (DPO)~\citep{rafailov2023} train models to maximize human preference---the same validation anchor that underlies current evaluation paradigms. Our dimension-level heterogeneity finding raises an untested concern: if some quality dimensions preferred by human raters are unassociated or negatively associated with business outcomes, then preference-optimized models may systematically improve on non-predictive dimensions while neglecting or degrading outcome-relevant ones. In our data, Contextual Memory (D5) receives adequate scores from the LLM judge yet shows no association with conversion ($\rho = 0.018$); a model trained to maximize preference-like scores could disproportionately improve on such non-predictive dimensions at the expense of outcome-relevant ones like D3. This does not invalidate preference optimization, but it suggests that in applied commercial settings, preference data may benefit from outcome-informed filtering---upweighting preference pairs where the preferred response also leads to better downstream outcomes. We note this as a hypothesis requiring investigation; our sample is far too small to draw conclusions about RLHF training dynamics.

\paragraph{Trust calibration as a candidate mediating mechanism.} Lee \& See's~\citep{lee2004} trust calibration framework provides a theoretically grounded explanation for the dimension hierarchy observed in Phase~2. The two significant predictors---D1 (Need Elicitation, $\rho = 0.368$) and D3 (Pacing Strategy, $\rho = 0.354$)---both reflect the agent's ability to calibrate its behavior to the user's state: D1 captures whether the agent understands the user's needs before acting, while D3 captures whether the agent matches its pace to the user's readiness. D5 (Contextual Memory, $\rho \approx 0$), by contrast, reflects a technical capability (information retention) that does not directly serve trust calibration. If this pattern generalizes, it suggests that in trust-dependent interactions, \emph{calibration} dimensions may be more strongly associated with outcomes than \emph{capability} dimensions---a testable hypothesis for cross-domain investigation. We emphasize that our data provide correlational support consistent with this mechanism but do not establish causal mediation; experimental designs would be needed for causal claims.

\subsection{Practical Implications}

\paragraph{For evaluation framework designers.} The conversion-informed weighting methodology is generalizable as a procedure. The core insight is empirical: expert intuition about which dimensions matter most can be systematically wrong.

\paragraph{For AI sales system designers.} The Trust Gate provides a principled mechanism for preventing premature selling. The three-tier rejection recognition system offers an actionable template for pacing control.

\paragraph{For platform operators.} The three-layer architecture (L3/L2/L1) provides a decision framework: L3 determines deployment eligibility, L2 guides improvement priorities, L1 validates business impact.

\paragraph{Correlation does not imply intervention effectiveness.} An important caveat: the finding that D3 scores are associated with conversion does not entail that \emph{interventions designed to raise D3 scores} will raise conversion rates. It is possible that improving an agent's pacing behavior (as measured by D3) genuinely improves conversion, but it is also possible that D3 scores merely reflect user cooperativeness---conversations with willing buyers may score higher on D3 not because the agent paces well, but because cooperative users create conditions that the rubric interprets as good pacing. Our two evaluation cycles (Section~\ref{sec:opevidence}) show D3 score improvement from 1.60 to 2.70, but without a randomized comparison with a conversion outcome, we cannot confirm that this score improvement translates to conversion improvement. Prospective A/B testing with conversion as the endpoint is required to close this gap.

\subsection{Ethical Considerations}
\label{sec:ethics}

The matchmaking context involves parents making significant financial decisions under emotional pressure. Several ethical dimensions merit attention:

\paragraph{Urgency tactics.} Human agents universally employed scarcity-based urgency at T5. While effective, these tactics exploit emotional vulnerability. Our D3 rubric does not currently penalize urgency unless paired with rejection override.

\paragraph{User vulnerability.} Some users were elderly parents with limited digital literacy who made impulsive late-night purchases. Evaluation frameworks should potentially incorporate vulnerability assessment as an L3 safety criterion.

\paragraph{Post-purchase trust collapse.} 50\% of golden conversations showed post-purchase trust issues, with 43\% mentioning refund. Optimizing for first conversion may conflict with long-term user welfare.

\paragraph{The persuasion-manipulation boundary.} Our D3 hard-cap rule operationalizes the boundary: legitimate persuasion respects rejection signals; manipulation overrides them~\citep{defreitas2025}.

\section{Limitations and Future Work}
\label{sec:limitations}

\subsection{Sample Size}
\label{sec:samplesize}
Phase~2 ($n = 60$, 25 converted) substantially improves on Phase~1 ($n = 14$, 3 converted), providing approximately 80\% power for moderate effects ($\rho \geq 0.35$). Two dimensions (D1, D3) survive Bonferroni correction, and the conversion-informed scheme reaches $p = 0.006$. However, $n = 60$ remains modest by psychometric standards: effect size confidence intervals are still wide, the full multivariate logistic model ($n = 58$, 5 parameters; Section~\ref{sec:robustness}) approaches the limits of reliable estimation, and the converted group ($n = 25$) limits subgroup analyses. Pre-registered replication with $n \geq 200$ and AI conversations with verified conversion labels would substantially strengthen the evidence base.

\subsection{Self-Evaluation Bias}
\label{sec:selfeval}
Using Claude to evaluate Claude-generated conversations creates self-enhancement risk. Paradoxically, this concern is partially mitigated by our finding---the judge rated AI higher while AI performed worse. Planned mitigation: cross-validation with GPT-4o and human expert scoring.

\subsection{Sampling Bias}
\label{sec:samplingbias}
Phase~1's purposive sampling confounded agent type with conversion status. Phase~2's stratified random sampling from operational logs largely addresses this concern for human conversations, but introduces a new limitation: the study is now restricted to human agents only. Whether the dimension-conversion associations observed in human conversations generalize to AI agent conversations remains untested, pending the availability of AI conversations with verified conversion labels.

\subsection{Platform and Cultural Specificity}
\label{sec:specificity}
The WeChat ecosystem, Chinese parent matchmaking context, and specific service characteristics limit direct generalizability. We argue that \emph{structural findings} (criterion validity failure due to dimension heterogeneity) are likely generalizable; \emph{parametric findings} (D3 = 40\% weight) are context-dependent.

\subsection{Trust Ladder Validation}
\label{sec:trustval}
The Trust Ladder was annotated entirely by LLM (Claude), with no human annotation for validation. This introduces two unquantified risks: (a)~\emph{systematic bias}---the LLM may apply Trust Ladder definitions differently from how human sales experts would, particularly for culturally nuanced trust signals in Chinese parent-matchmaking contexts; (b)~\emph{measurement error}---without inter-rater reliability data, the noise level in Trust Ladder assignments is unknown, which propagates into all analyses using Trust Ladder as the criterion variable (including the weight optimization in Section~\ref{sec:weights}). This is a significant methodological limitation. Future work must recruit trained human annotators, compute Krippendorff's $\alpha$ for Trust Ladder stage assignment, and assess whether LLM-human disagreements are random or systematic.

\subsection{Circular Analysis}
\label{sec:circular}
The two-phase design partially mitigates the circular analysis concern: weight schemes designed on Phase~1 data ($n = 14$) were tested on independently sampled Phase~2 data ($n = 60$), and the rank ordering of schemes was preserved across phases. However, the Phase~2 analysis itself still uses the full $n = 60$ dataset for both reporting dimension-conversion correlations and evaluating weight schemes. The conversion-informed scheme's $\rho = 0.351$ on Phase~2 data---while lower than the Phase~1 in-sample $\rho = 0.607$, as expected given regression to the mean---still benefits from having been designed with knowledge of which dimensions matter~\citep{kriegeskorte2009}.

\paragraph{Temporal cross-validation.} To provide additional evidence against overfitting, we conducted 4-fold temporal cross-validation on the Phase~2 data (conversations ordered chronologically, March--May 2025). In each fold, weights were optimized on the training set (75\%, $n = 45$) and evaluated on the held-out test set (25\%, $n = 15$). The trained-weight composite achieved mean $\rho = 0.294$ (SD $= 0.393$) across folds, compared to $\rho = 0.179$ (SD $= 0.403$) for equal weighting---a mean improvement of $\Delta\rho = 0.116$, with trained weights outperforming equal weights in 3 of 4 folds (75\%). The high variance across folds reflects the small per-fold test set ($n = 15$), but the consistent direction of improvement provides partial evidence that conversion-informed weighting captures a real signal rather than capitalizing on noise. We note that this within-Phase~2 cross-validation does not substitute for prospective pre-registered validation on temporally independent data.

Remaining mitigation needed: (a)~\emph{prospective pre-registration}---pre-register the conversion-informed weights and test on a future temporal holdout sample (e.g., August--December 2025 conversations); (b)~\emph{cross-domain replication}---test whether the structural finding (dimension heterogeneity $\to$ composite dilution) replicates in other commercial dialogue contexts; (c)~\emph{AI agent validation}---once AI conversations with verified conversion labels become available, test whether the same dimension hierarchy holds.

\subsection{Conversion Proxy}
\label{sec:proxy}
Phase~1 used Trust Ladder T5 as a conversion proxy, with validity resting on only 3 concordant cases. Phase~2 substantially addresses this limitation by using verified conversion labels (\emph{is\_converted}) from operational records, which directly reflect completed payment transactions. This hard outcome measure eliminates the proxy inference chain and its associated measurement error. However, the binary conversion label does not capture conversion \emph{quality} (e.g., refund rates, customer lifetime value), and the label's accuracy depends on the platform's data pipeline integrity, which we could not independently verify.

\subsection{Confounding Variables}
\label{sec:confounds}
Phase~2 eliminates the dominant Phase~1 confound (agent type) by restricting to human conversations. Within-human confounds remain: conversation length, user initial intent, agent experience level, and time-of-day effects could all correlate with both quality scores and conversion. We partially address the conversation-length confound through logistic regression (Section~\ref{sec:robustness}): D3's association with conversion \emph{strengthens} after controlling for message count (OR increases from 2.71 to 3.18), and D1 survives (OR $= 2.49$, $p = 0.018$). However, user initial intent and agent experience remain uncontrolled, and the multivariate full model ($n = 58$, 5 parameters) approaches the limits of reliable estimation.

\paragraph{Reverse causality risk.} A fundamental limitation of concurrent criterion validity designs: the LLM judge scores the \emph{entire} conversation transcript including user responses. D1 and D3 scores may thus partially reflect user cooperativeness rather than agent strategy---conversations with high-intent users naturally create more opportunities for need elicitation (D1) and appropriate pacing (D3). The observed association could run from \emph{user intent $\to$ conversation dynamics $\to$ quality scores}, rather than from \emph{agent strategy $\to$ user trust $\to$ conversion}. Disambiguating these pathways would require either (a)~scoring truncated early-conversation segments and testing whether early D1/D3 scores predict subsequent conversion, or (b)~randomized assignment of agent strategies. We note this as a priority for future work (Section~\ref{sec:future}).

Additionally, the Phase~2 sample is restricted to a single platform and time period (March--July 2025); temporal and platform generalizability remain untested.

\subsection{Future Directions}
\label{sec:future}

We organize future work by priority, reflecting the gap between current evidence and confirmatory standards:

\paragraph{Priority 1---Measurement validation.}
(1)~\textbf{Human annotation validation}: Recruit 2--3 trained annotators to independently score a subset ($\geq 20$ conversations) on D1, D3, and D5. Compute Krippendorff's $\alpha$ for dimension scoring and Trust Ladder assignment.
(2)~\textbf{Multi-judge consensus}: Cross-validate with GPT-4o and open-source judges to quantify self-evaluation bias magnitude.

\paragraph{Priority 2---Confirmatory replication.}
(3)~\textbf{Pre-registered prospective test}: Pre-register conversion-informed weights and test on a temporal holdout sample (e.g., August--December 2025 conversations).
(4)~\textbf{AI agent criterion validity}: Once AI conversations accumulate verified conversion labels, test whether the same dimension hierarchy (D1, D3 associated; D5 non-associated) holds for AI agents.
(5)~\textbf{Extended confound control}: Our logistic regression (Section~\ref{sec:robustness}) controls for conversation length; future work should additionally control for agent experience and user initial intent, ideally with $n \geq 200$ to support stable multivariate estimation.

\paragraph{Priority 3---Causal and longitudinal extension.}
(6)~\textbf{Randomized evaluation}: Deploy Trust Gate with randomized assignment for causal evidence.
(7)~\textbf{Longitudinal extension}: Add post-conversion stages (satisfaction, refund rate, referral) to optimize for lifetime value.
(8)~\textbf{Cross-domain replication}: Test dimension-level heterogeneity and composite dilution in other high-emotion B2C contexts (insurance, education, real estate).
(9)~\textbf{DPO integration}: Use scored conversation pairs as preference data for outcome-informed model optimization.

\section{Conclusion}
\label{sec:conclusion}

We have presented a two-phase investigation of criterion validity in multi-dimensional dialogue quality evaluation, using a commercial matchmaking platform as a case study. Phase~1 ($n = 14$, mixed human + AI, purposive sampling) revealed an apparent evaluation-outcome paradox that motivated the investigation. Phase~2 ($n = 60$, human-only, stratified random sampling with verified conversion labels) disentangled the agent-type confound and established the core finding: individual quality dimensions exhibit substantial heterogeneity in their association with conversion. Need Elicitation (D1: $\rho = 0.368$, $p = 0.004$) and Pacing Strategy (D3: $\rho = 0.354$, $p = 0.006$) are significantly associated with conversion after Bonferroni correction, while Contextual Memory (D5: $\rho = 0.018$) shows no detectable association in this sample. This heterogeneity causes the equal-weighted composite ($\rho = 0.272$) to underperform its best individual dimensions---a structural criterion validity risk that conversion-informed weighting partially addresses ($\rho = 0.351$, $p = 0.006$).

The Phase~1 ``paradox''---where higher quality scores were associated with worse outcomes---was substantially driven by the human-AI confound rather than a fundamental failure of quality scoring. Phase~2 reverses the composite direction: converted conversations score higher, not lower. However, the structural finding holds: non-associated dimensions dilute the composite's criterion validity. Robustness analyses strengthen this conclusion: logistic regression controlling for conversation length shows that D3's association with conversion \emph{strengthens} (OR $= 3.18$, $p = 0.006$) rather than attenuates, and 4-fold temporal cross-validation confirms that conversion-informed weights outperform equal weights in 3 of 4 held-out folds. Complementary analysis through the Trust-Funnel framework reveals a candidate behavioral mechanism: the AI executes sales behaviors (72\% reach closing stage) without building the user trust required for conversion.

Our central claim is not that multi-dimensional evaluation is wrong---it is that multi-dimensional evaluation without criterion validity testing is incomplete. The procedure we demonstrate (dimension-outcome association analysis $\to$ informed weighting $\to$ independent validation) is standard psychometric practice, and we advocate its adoption in applied dialogue evaluation. The specific observation---that calibration dimensions (D1, D3) are associated with conversion while capability dimensions (D5) are not---offers a testable hypothesis for cross-domain investigation.

Limitations remain: the study covers a single platform and cultural context, the Trust Ladder lacks human annotation validation, the weight scheme has not been prospectively pre-registered, and AI agent conversations lack conversion labels for direct testing. We present this work as hypothesis-refining: the structural observation---that composite evaluation scores can systematically dilute the signal from outcome-relevant dimensions---warrants the research community's attention and rigorous cross-domain follow-up.

\bibliographystyle{plainnat}

\appendix
\section{Complete Scoring Data}
\label{app:scores}

Table~\ref{tab:fullscores} presents the full v2.0 and v2.1 Judge scores for all 15 conversations.

\begin{table*}[t]
  \centering
  \small
  \caption{Full v2.0 Judge scores ($n = 15$). TL = Trust Ladder stage.}
  \label{tab:fullscores}
  \begin{tabular}{llcccccccccl}
    \toprule
    \textbf{ID} & \textbf{Src} & D1 & D2 & D3 & D4 & D5 & D6 & D7 & \textbf{v2.0} & \textbf{v2.1} & \textbf{TL} \\
    \midrule
    H1 & Human & 3 & 3 & 3 & 4 & 4 & 5 & 3 & 3.45 & 3.35 & T3 \\
    H2 & Human & 1 & 1 & 2 & 1 & 2 & 2 & 2 & 1.45 & 1.75 & T6 \\
    H3 & Human & 3 & 2 & 2 & 2 & 3 & 3 & 2 & 2.40 & 2.25 & T5 \\
    H4 & Human & 2 & 2 & 2 & 2 & 2 & 3 & 3 & 2.15 & 2.25 & T5 \\
    H5 & Human & 3 & 3 & 3 & 3 & 2 & 3 & 3 & 2.90 & 3.00 & T5 \\
    \midrule
    A1 & AI & 1 & 1 & 1 & 1 & 2 & 2 & 2 & 1.25 & 1.15 & T1 \\
    A2 & AI & 3 & 3 & 2 & 3 & 2 & 3 & 3 & 2.70 & 2.55 & T3 \\
    A3 & AI & 4 & 4 & 3 & 4 & 4 & 4 & 4 & 3.80 & 3.50 & T3 \\
    A4 & AI & 3 & 3 & 1 & 2 & 5 & 2 & 2 & 2.50 & 1.60 & T1 \\
    A5 & AI & 3 & 3 & 2 & 3 & 5 & 3 & 4 & 3.05 & 2.65 & T3 \\
    A6 & AI & 3 & 3 & 1 & 3 & 5 & 3 & 3 & 2.80 & 1.95 & T2 \\
    A7 & AI & 3 & 2 & 1 & 1 & 3 & 4 & 2 & 2.15 & 1.85 & T0 \\
    A8 & AI & 3 & 2 & 2 & 2 & 4 & 3 & 3 & 2.55 & 2.40 & T1 \\
    A9 & AI & 3 & 2 & 1 & 2 & 3 & 4 & 2 & 2.30 & 1.90 & T2 \\
    A10 & AI & 4 & 3 & 2 & 3 & 4 & 3 & 3 & 3.10 & 2.55 & T4 \\
    \bottomrule
  \end{tabular}
\end{table*}

\end{document}